\documentclass[twoside,11pt]{article}
\usepackage{times} 
\usepackage{theapa}
\usepackage{amsmath}
\usepackage{amsthm}
\usepackage{amssymb}
\usepackage{epsf}
\usepackage{glossary}
\usepackage{jair}                                                               
\usepackage[dvips]{graphicx}
\usepackage{latexsym}
\usepackage{xspace}

\usepackage{algorithm}
\usepackage{algorithmic}

\renewcommand{\(}{\left(}
\renewcommand{\)}{\right)}

\renewcommand{\[}{\left[}
\renewcommand{\]}{\right]}

\newcommand{\Jt}{{\tilde{J}}}

\newcommand{\rbar}{{\bar{r}}}
\newcommand{\Jbar}{{\bar{J}}}

\newcommand{\ie}{{\em i.e.\ }}

\newcommand{\iid}{{i.i.d.\ }}

\newcommand{\B}{{\cal B}}

\renewcommand{\P}{{\cal P}}

\renewcommand{\S}{{\cal S}}
\newcommand{\T}{{\cal T}}
\newcommand{\U}{{\cal U}}

\newcommand{\Y}{{\cal Y}}

\renewcommand{\P}{{\cal P}}

\DeclareMathOperator{\argmin}{argmin}

\DeclareMathOperator{\diag}{diag}

\DeclareMathOperator{\TD}{TD}

\DeclareMathOperator{\VAPS}{VAPS}
\DeclareMathOperator{\DES}{DES}

\DeclareMathOperator{\MCG}{MCG}
\DeclareMathOperator{\GPOMDP}{GPOMDP}

\DeclareMathOperator{\REINFORCE}{REINFORCE}
\DeclareMathOperator{\MDP}{MDP}
\DeclareMathOperator{\POMDP}{POMDP}

\newcommand{\mcg}{{\mbox{\small$\MCG$}}}
\newcommand{\vaps}{{\mbox{\small$\VAPS$}}}

\newcommand{\des}{{\mbox{\small$\DES$}}}

\newcommand{\pomdpg}{{\mbox{\small$\GPOMDP$}}}
\newcommand{\gpomdp}{{\mbox{\small$\GPOMDP$}}}

\newcommand{\reinforce}{{\mbox{\small$\REINFORCE$}}}
\newcommand{\mdp}{{\mbox{\small$\MDP$}}}
\newcommand{\mdps}{{\mbox{\small$\MDP$}}s}
\newcommand{\pomdp}{{\mbox{\small$\POMDP$}}}
\newcommand{\pomdps}{{\mbox{\small$\POMDP$}}s}

\newcommand{\nb}{{\nabla_{\!\!\beta}}}
\newcommand{\Expect}{\mathbold{E}}

\newcommand{\E}{\Expect}
\newcommand{\mathbold}[1]{\mbox{\boldmath $\bf#1$}}

\newcommand{\R}{{\mathbb R}}

\usepackage{color}

\definecolor{darkred}{rgb}{0.7,0.2,0.2}

\definecolor{bgblue}{rgb}{0.04,0.39,0.53}

\newtheorem{theorem}{Theorem}
\newtheorem{proposition}[theorem]{Proposition}
\newtheorem{assumption}{Assumption}
\newtheorem{lemma}[theorem]{Lemma}

\begin{document}

\title{Infinite-Horizon Policy-Gradient Estimation}
\author{\name Jonathan Baxter
\email jbaxter@whizbang.com \\
\addr{WhizBang! Labs.} \\
\addr{4616 Henry Street Pittsburgh, PA  15213} \\
\name Peter L. Bartlett
\email bartlett@barnhilltechnologies.com\\
\addr{BIOwulf Technologies.} \\
\addr{2030 Addison Street, Suite 102, Berkeley, CA 94704}}
\jairheading{15}{2001}{}{9/00}{10/01}
\firstpageno{319}
\maketitle
\ShortHeadings{Policy-Gradient Estimation}{Baxter and Bartlett}
\begin{abstract}
Gradient-based approaches to direct policy search in reinforcement learning
have received much recent attention as a means to solve problems of
partial observability and to avoid some of the problems associated
with policy degradation in value-function methods.  In this paper
we introduce \gpomdp, a simulation-based algorithm for generating a
{\em biased} estimate of the gradient of the {\em average reward} in
Partially Observable Markov Decision Processes (\pomdps) controlled by
parameterized stochastic policies. A similar algorithm was proposed by
\citeA{kimura95}. The algorithm's chief advantages are that it requires
storage of only twice the number of policy parameters, uses one free
parameter $\beta\in [0,1)$ (which has a natural interpretation in terms
of bias-variance trade-off), and requires no knowledge of the underlying
state. We prove convergence of \pomdpg, and show how the correct
choice of the parameter $\beta$ is related to the {\em mixing time}
of the controlled \pomdp. We briefly describe extensions of \pomdpg\
to controlled Markov chains, continuous state, observation and control
spaces, multiple-agents,
higher-order derivatives, and a version for training stochastic policies
with internal states.  In a companion paper \cite{jair_01b} we show how the
gradient estimates generated by \gpomdp\ can be used in both a traditional
stochastic gradient algorithm and a conjugate-gradient procedure to find
local optima of the average reward.

\end{abstract}

\section{Introduction}
\label{section:intro}
Dynamic Programming is the method of choice for solving problems
of decision making under uncertainty \cite{bertsekas95}. However,
the application of Dynamic Programming becomes problematic in large or 
infinite state-spaces, in situations where the system dynamics are
unknown, or when the state is only partially observed. In such cases
one looks for approximate techniques that rely on simulation, rather
than an explicit model, and parametric representations of either the
value-function or the policy, rather than exact representations. 

Simulation-based methods that rely on a parametric form of the value
function tend to go by the name ``Reinforcement Learning,'' and have
been extensively studied in the Machine Learning literature
\cite{bertsekas96,sutton98}. This approach has yielded some remarkable
empirical successes in a number of different domains, including
learning to play checkers \cite{samuel59}, backgammon
\cite{tesauro92,tesauro94}, and chess
\cite{ml_00a}, job-shop scheduling \cite{zhang95} and dynamic 
channel allocation \cite{singh97}. 

Despite this success, most algorithms for training approximate value
functions suffer from the same theoretical flaw: the performance
of the greedy policy derived from the approximate value-function is
not guaranteed to improve on each iteration, and in fact can be worse
than the old policy by an amount equal to the {\em maximum}
approximation error over all states. This can happen even when the
parametric class contains a value function whose corresponding greedy
policy is optimal. We illustrate this with a concrete and very simple
example in Appendix~\ref{section:badshit}.

An alternative approach that circumvents this problem---the approach
we pursue here---is to consider a class of {\em stochastic policies}
parameterized by $\theta \in \R^K$, compute the gradient with respect
to $\theta$ of the average reward, and then improve the policy by
adjusting the parameters in the gradient direction. Note that the
policy could be directly parameterized, or it could be generated
indirectly from a value function. In the latter case the
value-function parameters are the parameters of the policy, but
instead of being adjusted to minimize error between the approximate
and true value function, the parameters are adjusted to directly
improve the performance of the policy generated by the value function.

These ``policy-gradient'' algorithms have a long history in 
Operations Research, Statistics, Control Theory, Discrete Event
Systems and Machine Learning. Before describing the contribution of
the present paper, it seems appropriate to introduce some background
material explaining this approach. Readers already familiar with this
material may want to skip directly to section \ref{section:us}, where the
contributions of the present paper are described. 

\subsection{A Brief History of Policy-Gradient Algorithms}
\label{subsec:history}
For large-scale problems or problems where the system dynamics are
unknown, the performance gradient will not be computable in closed
form\footnote{See equation \eqref{eq:gradeta} for a closed-form
expression for the performance gradient.}. Thus the challenging aspect
of the policy-gradient approach is to find an algorithm for estimating
the gradient via {\em simulation}.  Naively, the gradient can be
calculated numerically by adjusting each parameter in turn and
estimating the effect on performance via simulation (the so-called
{\em crude Monte-Carlo} technique), but that will be prohibitively
inefficient for most problems. Somewhat surprisingly, under mild
regularity conditions, it turns out that the full gradient can be
estimated from a {\em single} simulation of the system. The technique
is called the {\em score function} or {\em likelihood ratio} method
and appears to have been first proposed in the sixties
\cite{aleksandrov68,rubinstein69} for computing performance gradients
in \iid (independently and identically distributed)
processes. 

Specifically, suppose $r(X)$ is a performance function that depends on
some random variable $X$, and $q(\theta,x)$ is the probability that $X=x$,
parameterized by $\theta\in\R^K$.  Under mild regularity conditions,
the gradient with respect to $\theta$ of the expected performance,
\begin{equation}
\label{eq:expperf}
\eta(\theta) = \E r(X),
\end{equation}
may be written 
\begin{equation}
\label{eq:gradexpperf}
\nabla\eta(\theta) = \E r(X)
  \frac{\nabla q(\theta,X)}{q(\theta,X)}. 
\end{equation}
To see this, rewrite \eqref{eq:expperf} as a sum
$$
\eta(\theta) = \sum_x r(x) q(\theta,x),
$$
differentiate (one source of the requirement of
``mild regularity conditions'') to obtain
$$
\nabla\eta(\theta) = \sum_x r(x) \nabla q(\theta,x),
$$
rewrite as
$$
\nabla\eta(\theta) = \sum_x r(x) \frac{\nabla q(\theta,x)}{q(\theta, x)}
q(\theta, x),
$$
and observe that this formula is equivalent to
\eqref{eq:gradexpperf}. 

If a simulator is available
to generate samples $X$ distributed according to $q(\theta,x)$, then
any sequence $X_1,X_2,\dots,X_N$ generated \iid according to
$q(\theta,x)$ gives an unbiased estimate,
\begin{equation}
\label{eq:approxexpperf}
\hat{\nabla} \eta(\theta) = \frac1N\sum_{i=1}^N r(X_i) \frac{\nabla
q(\theta,X_i)}{q(\theta,X_i)},
\end{equation}
of $\nabla\eta(\theta)$. By the law of large numbers, $\hat{\nabla}
\eta(\theta) \rightarrow \nabla\eta(\theta)$ with probability one. 
The quantity $\nabla q(\theta,X) / q(\theta,X)$ is known as the
{\em likelihood ratio} or {\em score function} in classical
statistics. If the performance function $r(X)$ also depends on
$\theta$, then $r(X) \nabla q(\theta,X) / q(\theta,X)$ is replaced
by $\nabla r(\theta,X) + r(\theta, X)\nabla q(\theta,X) / q(\theta,X)$ in
\eqref{eq:gradexpperf}. 

\subsubsection{Unbiased Estimates of the Performance Gradient for
Regenerative Processes} 
\label{section:regen}

Extensions of the likelihood-ratio method to {\em regenerative
processes} (including Markov Decision Processes or \mdps) were given
by \citeA{glynn86,glynn90,glynn95} and
\citeA{reiman86,reiman89}, and independently for {\em episodic} Partially Observable
Markov Decision Processes (\pomdps) by \citeA{williams92}, who
introduced the \reinforce\ algorithm\footnote{A {\em thresholded}
version of these algorithms for neuron-like elements was described
earlier in \citeA{BarSutAnd83}.}. Here the \iid samples $X$ of the
previous section are {\em sequences} of states $X_0,\dots,X_T$ (of
random length) encountered between visits to some designated recurrent
state $i^*$, or sequences of states from some start state to a goal
state. In this case $\nabla q(\theta,X) / q(\theta,X)$ can be written
as a sum
\begin{equation}
\label{eq:mdpgrad}
\frac{\nabla q(\theta,X)}{q(\theta,X)} = \sum_{t=0}^{T-1} \frac{\nabla
p_{X_t X_{t+1}}(\theta)}{p_{X_t X_{t+1}}(\theta)}, 
\end{equation}
where $p_{X_t X_{t+1}}(\theta)$ is the transition probability from
$X_t$ to $X_{t+1}$ given parameters $\theta$. Equation
\eqref{eq:mdpgrad} admits a recursive
computation over the course of a regenerative cycle of the form 
$z_0=0\in\R^K$, and after each state transition $X_t\rightarrow X_{t+1}$, 
\begin{equation}
\label{eq:z}
z_{t+1} = z_t + \frac{\nabla p_{X_t X_{t+1}}(\theta)}{p_{X_t X_{t+1}}(\theta)},
\end{equation}
so that each term $r(X)\nabla q(\theta,X) / q(\theta,X)$ in the
estimate \eqref{eq:approxexpperf} is of the form\footnote{The vector
$z_T$ is known in reinforcement learning as an {\em eligibility
trace}. This terminology is used in \citeA{BarSutAnd83}.}
$r(X_0,\dots, X_T) z_T$. If, in addition, $r(X_0,\dots, X_T)$ can be
recursively computed by
  $$
    r(X_0, \dots, X_{t+1}) = \phi(r(X_0, \dots, X_{t}), X_{t+1})
  $$
for some function $\phi$, then the estimate
$r(X_0,\dots, X_T) z_T$ for each cycle can be computed using storage
of only $K+1$ parameters ($K$ for $z_t$ and $1$ parameter to update
the performance function $r$). Hence, the entire estimate \eqref{eq:approxexpperf} can
be computed with storage of only $2 K + 1$ real parameters, as
follows.
\paragraph{Algorithm 1.1: Policy-Gradient Algorithm for Regenerative Processes.}
\begin{enumerate}
\item 
Set $j=0$, $r_0 = 0$, $z_0 = 0$, and $\Delta_0 = 0$ ($z_0, \Delta_0
\in\R^K$).
\item 
\label{loop}
For each state transition $X_t \rightarrow X_{t+1}$: 
        \begin{itemize}
        \item If the episode is finished (that is, $X_{t+1} = i^*)$, set \\
        $
        \Delta_{j+1} = \Delta_j + r_t z_t$, \\
        $j = j+1$,\\ 
        $z_{t+1}= 0$,\\
        $r_{t+1} = 0$.  
        \item Otherwise, set \\
        $
        z_{t+1} = z_t + \frac{\nabla p_{X_t
        X_{t+1}}(\theta)}{p_{X_t X_{t+1}}(\theta)},
        $\\
        $r_{t+1} = \phi(r_t, X_{t+1})$. 
        \end{itemize}
\item If $j = N$ return $\Delta_N / N$, otherwise goto \ref{loop}. 
\end{enumerate}
Examples of recursive performance functions include the sum of a
scalar reward over a cycle, $r(X_0,\dots, X_T) = \sum_{t=0}^T r(X_t)$
where $r(i)$ is a scalar reward associated with state $i$ (this
corresponds to $\eta(\theta)$ being the {\em average reward}
multiplied by the expected recurrence time $\E_\theta\[T\]$); the
negative length of the cycle (which can be implemented by assigning a
reward of $-1$ to each state, and is used when the task is to mimimize
time taken to get to a goal state, since $\eta(\theta)$ in this case
is just $-\E_\theta\[T\]$); the {\em discounted reward} from
the start state, $r(X_0, \dots, X_T) = \sum_{t=0}^T \alpha^t r(X_t)$,
where $\alpha\in [0,1)$ is the discount factor, and so on.

As \citeA{williams92} pointed out, a further simplification is
possible in the case that $r_T = r(X_0,\dots, X_T)$ is a sum of scalar
rewards $r(X_t, t)$ depending on the state and possibly the time $t$
since the starting state (such as $r(X_t,t) = r(X_t)$, or $r(X_t,t) =
\alpha^t r(X_t)$ as above). In that case, the update $\Delta$  from a
single regenerative cycle may be written as 
$$
\Delta = \sum_{t=0}^{T-1}
  \frac{\nabla p_{X_t X_{t+1}}(\theta)}{p_{X_t X_{t+1}}(\theta)}
\[\sum_{s=0}^t r(X_s,s) + \sum_{s=t+1}^T r(X_s,s)\]. 
$$ 
Because changes in $p_{X_tX_{t+1}}(\theta)$ have no influence on
the rewards $r(X_s, s)$ associated with earlier states ($s\leq t$), we
should be able to drop the first term in the parentheses on the
right-hand-side and write
\begin{equation}
\label{eq:altdelta}
\Delta = \sum_{t=0}^{T-1} \frac{\nabla p_{X_t X_{t+1}}(\theta)}{p_{X_t
X_{t+1}}(\theta)} \sum_{s=t+1}^T r(X_s,s).
\end{equation}
Although the proof is not entirely trivial, this intuition can
indeed be shown to be correct. 

Equation \eqref{eq:altdelta} allows an even simpler recursive formula
for estimating the performance gradient. Set $z_0=\Delta_0 = 0$, and
introduce a new variable $s=0$. As before, set $z_{t+1} = z_t +\nabla
p_{X_t X_{t+1}}(\theta) / p_{X_t X_{t+1}}(\theta)$ and $s=s+1$ if
$X_{t+1} \neq i^*$, or $s = 0$ and $z_{t+1} = 0$ otherwise. But now,
on {\em each iteration}, set $\Delta_{t+1} = r(X_t,s) z_t +
\Delta_t$. Then $\Delta_t/t$ is our estimate of
$\nabla\eta(\theta)$. Since $\Delta_t$ is updated on every iteration,
this suggests that we can do away with $\Delta_t$ altogether and
simply update $\theta$ directly: $\theta_{t+1} = \theta_t + \gamma_t
r(X_t,s) z_t$, where the $\gamma_t$ are suitable
step-sizes\footnote{The usual requirements on $\gamma_t$ for
convergence of a stochastic gradient algorithm are $\gamma_t > 0$,
$\sum_{t=0}^\infty \gamma_t =
\infty$, and $\sum_{t=0}^\infty \gamma_t^2 < \infty$.}. Proving
convergence of such an algorithm is not as straightforward as normal
stochastic gradient algorithms because the updates $r(X_t) z_t$ are
not in the gradient direction (in expectation), although the
sum of these updates over a regenerative cycle are. \citeA{marbach98}
provide the only convergence proof that we know of, albeit for a slightly
different update of the form $\theta_{t+1} = \theta_t + \gamma_t
\[r(X_t,s) - \hat{\eta}(\theta_t)\] z_t$, where $\hat{\eta}(\theta_t)$
is a moving
estimate of the expected performance, and is also updated on-line
(this update was first suggested in the context of \pomdps\
by \shortciteA{jaakola95}).

\citeA{marbach98} also considered the case of
$\theta$-dependent rewards (recall the discussion after
\eqref{eq:approxexpperf}), as did \citeA{baird98} with their ``\vaps''  
algorithm ({\em Value And Policy Search}). This last paper contains an
interesting insight: through suitable choices of the performance
function $r(X_0,\dots, X_T, \theta)$, one can combine policy-gradient search
with approximate value function methods.  The resulting algorithms can
be viewed as {\em actor-critic} techniques in the spirit of
\citeA{BarSutAnd83}; the policy is the {\em actor} and the value
function is the {\em critic}. The primary motivation is to reduce
variance in the
policy-gradient estimates. Experimental evidence for this phenomenon
has been presented by a number of authors, including
\citeA{BarSutAnd83}, \citeA{kimura98a}, and \citeA{baird98}.
More recent work on this
subject includes that of \shortciteA{sutton99} and \shortciteA{konda99}. We
discuss the use of \vaps-style updates further in Section
\ref{section:pdr}.

So far we have not addressed the question of how the parameterized
state-transition probabilities $p_{X_tX_{t+1}}(\theta)$ arise. Of
course, they could simply be generated by parameterizing the matrix of
transition probabilities directly. Alternatively, in the case of
\mdps\ or \pomdps, state transitions are typically generated by
feeding an {\em observation} $Y_t$ that depends stochastically on the
state $X_t$ into a parameterized {\em stochastic policy}, which
selects a {\em control} $U_t$ at random from a set of available
controls (approximate value-function based approaches that generate
controls stochastically via some form of lookahead also fall into this
category). The distribution over successor states
$p_{X_tX_{t+1}}(U_t)$ is then a fixed function of the control. If we
denote the probability of control $u_t$ given parameters $\theta$ and 
observation $y_t$ by $\mu_{u_t}(\theta,y_t)$, then all of the above
discussion carries through with $\nabla p_{X_tX_{t+1}}(\theta)/p_{X_t
X_{t+1}}(\theta)$ replaced by
$\nabla\mu_{U_t}(\theta,Y_t)/\mu_{U_t}(\theta,Y_t)$. In that case,
Algorithm 1.1 is precisely Williams' \reinforce\ algorithm.

Algorithm 1.1 and the variants
above have been extended to cover multiple agents
\shortcite{peshkin00}, policies with internal state
\shortcite{meuleau99}, and importance sampling methods
\shortcite{meuleau00}. We also refer the reader to
the work of \citeA{rubinstein93} and \citeA{rubinstein98}
for in-depth analysis of the
application of the likelihood-ratio method to Discrete-Event Systems
(\des), in particular networks of queues. Also worth mentioning is the
large literature on Infinitesimal Perturbation Analysis (IPA), which
seeks a similar goal of estimating performance gradients, but operates
under more restrictive assumptions than the likelihood-ratio approach;
see, for example, \citeA{ho91}.

\subsubsection{Biased Estimates of the Performance Gradient}
 
All the algorithms described in the previous section rely on an
identifiable recurrent state $i^*$, either to update the gradient estimate,
or in the case of the on-line algorithm, to zero the eligibility trace
$z$. This reliance on a recurrent state can be problematic for two main
reasons: 
\begin{enumerate} 
\item 
\label{p1}
The {\em variance} of the algorithms is related to the recurrence time
between visits to $i^*$, which will typically grow as the state space
grows. Furthermore, the time between visits depends on the parameters
of the policy, and states that are frequently visited for the initial
value of the parameters may become very rare as performance improves.
\item 
\label{p2}
In situations of {\em partial observability} it may be difficult to
estimate the underlying states, and therefore to determine when the
gradient estimate should be updated, or the eligibility trace zeroed.
\end{enumerate}

If the system is available only through simulation, it seems difficult
(if not impossible) to obtain {\em unbiased} estimates of the gradient
direction without access to a recurrent state. Thus, to solve \ref{p1}
and \ref{p2}, we must look to {\em biased} estimates. Two principle
techniques for introducing bias have been proposed, both of which may
be viewed as artificial truncations of the eligibility trace $z$. The
first method takes as a starting point the formula\footnote{For ease of
exposition, we have kept the expression for $z$ in terms of the likelihood
ratios $\nabla p_{X_s X_{s+1}}(\theta)/p_{X_s X_{s+1}}(\theta)$ which
rely on the availability of the underlying state $X_s$. If $X_s$ is not
available, $\nabla p_{X_s X_{s+1}}(\theta)/p_{X_s X_{s+1}}(\theta)$ should
be replaced with $\nabla\mu_{U_s}(\theta, Y_s)/\mu_{U_s}(\theta, Y_s)$.}
for the eligibility trace at time $t$:
  $$
    z_t = \sum_{s=0}^{t-1}
      \frac{\nabla p_{X_s X_{s+1}}(\theta)}{p_{X_s X_{s+1}}(\theta)}
  $$
and simply truncates it at some (fixed, not random) number of terms $n$
looking backwards \cite{glynn90,rubinstein91,rubinstein92,cao98}:
  \begin{equation}\label{eq:ztn}
z_t(n) := \sum_{s=t-n}^{t-1} \frac{\nabla p_{X_s X_{s+1}}(\theta)}{p_{X_s
X_{s+1}}(\theta)}.
  \end{equation}
The eligibility trace $z_t(n)$ is then updated after each transition
$X_t\rightarrow X_{t+1}$ by 
\begin{equation}
\label{eq:ztnupdate}
z_{t+1}(n) = z_t(n) + \frac{\nabla p_{X_t X_{t+1}}(\theta)}{p_{X_t
X_{t+1}}(\theta)} - \frac{\nabla p_{X_{t-n} X_{t-n+1}}(\theta)}{p_{X_{t-n}
X_{t-n+1}}(\theta)}, 
\end{equation}
and in the case of state-based rewards $r(X_t)$,
the estimated gradient direction after $T$ steps is
\begin{equation}
\label{eq:etan}
\hat{\nabla}_n\eta(\theta) := \frac1{T-n+1} \sum_{t=n}^T z_t(n)r(X_t). 
\end{equation}
Unless $n$ exceeds the maximum recurrence time (which is infinite in
an ergodic Markov chain), $\hat{\nabla}_n\eta(\theta)$ is a biased
estimate of the gradient direction, although as $n\rightarrow\infty$,
the bias approaches zero. However the {\em variance} of
$\hat{\nabla}_n\eta(\theta)$ diverges in the limit of large $n$. This
illustrates a natural trade-off in the selection of the parameter $n$:
it should be large enough to ensure the bias is acceptable
(the expectation of $\hat{\nabla}_n\eta(\theta)$ should at least be within $90^\circ$ of
the true gradient direction), but not so large that the variance is
prohibitive. Experimental results by
\citeA{cao98} illustrate nicely this bias/variance trade-off.

One potential difficulty with this method is that the likelihood
ratios $\nabla p_{X_s X_{s+1}}(\theta)/p_{X_s X_{s+1}}(\theta)$ must
be remembered for the previous $n$ time steps, requiring storage of $K
n$ parameters. Thus, to obtain small bias, the memory may have to grow
without bound. An alternative approach that requires a fixed amount of
memory is to {\em discount} the eligibility trace, rather than
truncating it: 
\begin{equation}
\label{eq:discupdate}
z_{t+1}(\beta) := \beta z_t(\beta) +
  \frac{\nabla p_{X_t X_{t+1}}(\theta)}{p_{X_t X_{t+1}}(\theta)}, 
\end{equation}
where $z_0(\beta) = 0$ and $\beta\in[0,1)$ is a discount factor. In
this case the estimated gradient direction after $T$ steps is simply
\begin{equation}
\label{eq:disc}
\hat{\nb}\eta(\theta) := \frac1T\sum_{t=0}^{T-1} r(X_t) z_t(\beta). 
\end{equation}
This is precisely the estimate we analyze in the present paper. A
similar estimate with $r(X_t) z_t(\beta)$ replaced by $(r(X_t) - b)
z_t(\beta)$ where $b$ is a {\em reward baseline} was proposed by
\shortciteA{kimura95,kimura97} and for continuous control by
\citeA{kimura98}.  In fact the use of $(r(X_t) - b)$ in place of
$r(X_t)$ does not affect the expectation of the estimates of the
algorithm (although judicious choice of the reward baseline $b$ can
reduce the variance of the estimates).  While the algorithm presented
by \citeA{kimura95} provides estimates of the expectation under the
stationary distribution of the gradient of the discounted reward, we
will show that these are in fact biased estimates of the gradient of
the expected discounted reward. This arises because the stationary
distribution itself depends on the parameters. A similar estimate to
\eqref{eq:disc} was also proposed by \citeA{marbach98}, but this time
with $r(X_t) z_t(\beta)$ replaced by $(r(X_t) - \hat{\eta}(\theta))
z_t(\beta)$, where $\hat{\eta}(\theta)$ is an estimate of the average
reward, and with $z_t$ zeroed on visits to an identifiable recurrent
state.

As a final note, observe that the eligibility traces $z_t(\beta)$ and
$z_t(n)$ defined by \eqref{eq:discupdate} and \eqref{eq:ztnupdate}
are simply {\em filtered} versions of the sequence $\nabla p_{X_t
X_{t+1}}(\theta)/p_{X_t X_{t+1}}(\theta)$, a first-order, infinite impulse
response filter in the case of $z_t(\beta)$ and an $n$-th order, finite
impulse response filter in the case of $z_t(n)$. This raises the question,
not addressed in this paper, of whether there is an interesting theory
of optimal filtering for policy-gradient estimators.

\subsection{Our Contribution}
\label{section:us}
We describe \gpomdp, a general algorithm based upon \eqref{eq:disc}
for generating a {\em biased} estimate of the performance gradient
$\nabla \eta(\theta)$ in general \pomdps\ controlled by parameterized
stochastic policies. Here $\eta(\theta)$ denotes the {\em average}
reward of the policy with parameters $\theta\in\R^K$.  \gpomdp\ does
not rely on access to an underlying recurrent state. Writing
$\nb\eta(\theta)$ for the expectation of the estimate produced by
\gpomdp, we show that $\lim_{\beta\rightarrow 1} \nb\eta(\theta) =
\nabla\eta(\theta)$, and more quantitatively that $\nb\eta(\theta)$ is
close to the true gradient provided $1/(1-\beta)$ exceeds the {\em
  mixing time} of the Markov chain induced by the \pomdp\footnote{The
  mixing-time result in this paper applies only to Markov chains with
  distinct eigenvalues. Better estimates of the bias and variance of
  \gpomdp\ may be found in \citeA{jcss_01}, for more general Markov
  chains than those treated here, and for more refined notions of the
  mixing time. Roughly speaking, the variance of \gpomdp\ grows with
  $1/(1-\beta)$, while the bias decreases as a function of
  $1/(1-\beta)$.}.  As with the truncated estimate above, the
trade-off preventing the setting of $\beta$ arbitrarily close to $1$
is that the variance of the algorithm's estimates increase as $\beta$
approaches $1$.  We prove convergence with probability 1 of \gpomdp\ 
for both discrete and continuous observation and control spaces. We
present algorithms for both general parameterized Markov chains and
\pomdps\ controlled by parameterized stochastic policies.

There are several extensions to \gpomdp\ that we have investigated
since the first version of this paper was written.  We outline these
developments briefly in Section \ref{section:newstuff}.

In a companion paper we show how the gradient estimates produced by
\gpomdp\ can be used to perform gradient ascent on the average reward
$\eta(\theta)$ \cite{jair_01b}. We describe both traditional
stochastic gradient algorithms, and a conjugate-gradient algorithm
that utilizes gradient estimates in a novel way to perform line
searches. Experimental results are presented illustrating both the
theoretical results of the present paper on a toy problem, and
practical aspects of the algorithms on a number of more realistic
problems.

\section{The Reinforcement Learning Problem}
\label{section:rl}

We model reinforcement learning as a Markov
decision process (\mdp) with a finite state space $\S = \{1,\dots,
n\}$, and a stochastic matrix\footnote{A {\em stochastic} matrix
$P=\[p_{ij}\]$ has $p_{ij} \geq 0$ for all $i,j$ and $\sum_{j=1}^n
p_{ij} = 1$ for all $i$.}  $P = \[p_{ij}\]$ giving the probability of
transition from state $i$ to state $j$. Each state $i$ has an
associated reward\footnote{All the results in the present paper
apply to bounded stochastic rewards, in which case $r(i)$ is the
expectation of the reward in state $i$.} $r(i)$. The matrix $P$ belongs to a
parameterized class of stochastic matrices, $\P:= \{P(\theta)
\colon \theta \in\R^K\}$.  Denote the Markov chain corresponding to
$P(\theta)$ by $M(\theta)$.  We assume that these Markov
chains and rewards satisfy the following assumptions:
\begin{assumption}
\label{ass:station}
\sloppy
Each $P(\theta) \in \P$ has a unique stationary distribution $\pi(\theta)
:= \[\pi(\theta, 1), \dots, \pi(\theta,n)\]'$ satisfying the {\em
balance equations}
\begin{equation}
\label{eq:balance}
\pi'(\theta) P(\theta) = \pi'(\theta)
\end{equation}
(throughout $\pi'$ denotes the transpose of $\pi$).
\end{assumption}
\begin{assumption}
\label{ass:rbound}
The magnitudes of the rewards, $|r(i)|$, are uniformly bounded by
$R < \infty$ for all states $i$.
\end{assumption}
Assumption \ref{ass:station} ensures that the Markov chain forms a
single recurrent class for all parameters $\theta$. Since any
finite-state Markov chain always ends up in a recurrent class, and it
is the properties of this class that determine the long-term average
reward, this assumption is mainly for convenience so that we do not
have to include the recurrence class as a quantifier in our theorems.
However, when we consider gradient-ascent algorithms \citeA{jair_01b},
this assumption becomes more
restrictive since it guarantees that the recurrence
class cannot change
as the parameters are adjusted.

Ordinarily, a discussion of \mdps\ would not be complete without some
mention of the actions available in each state and the space of
policies available to the learner. In particular, the parameters
$\theta$ would usually determine a policy (either directly or
indirectly via a value function), which would then determine the
transition probabilities $P(\theta)$. However, for our purposes we do
not care {\em how} the dependence of $P$ on $\theta$ arises, just that
it satisfies Assumption~\ref{ass:station} (and some differentiability
assumptions that we shall meet in the next section).  Note also that
it is easy to extend this setup to the case where the rewards
also depend on the parameters $\theta$ or on the transitions $i\to j$.
It is equally straightforward to extend our algorithms and results to these
cases. See Section~\ref{section:cdr} for an illustration.

The goal is to find a $\theta \in\R^K$ maximizing the {\em average reward}:
  $$
    \eta(\theta) := \lim_{T\rightarrow\infty}
      \E_\theta\[\left. \frac1T\sum_{t=0}^{T-1} r(X_t) \right|
        X_0 = i\],
  $$
where $\E_\theta$ denotes the expectation over all sequences $X_0, X_1,
\dots,$ with transitions generated according to $P(\theta)$.
Under Assumption~\ref{ass:station}, $\eta(\theta)$ is independent of
the starting state $i$ and is equal to
\begin{equation}
\label{eq:avgreward}
\eta(\theta) = \sum_{i=1}^n \pi(\theta,i) r(i) = \pi'(\theta) r,
\end{equation}
where $r = \[r(1), \dots, r(n)\]'$ \cite{bertsekas95}.

\section{Computing the Gradient of the Average Reward}
\label{sec:gradascent}

For general \mdps\ little will be known about the average reward
$\eta(\theta)$, hence finding its optimum will be problematic. However,
in this section we will see that under general assumptions the
gradient $\nabla\eta(\theta)$ exists, and so local optimization of
$\eta(\theta)$ is possible.


To ensure the existence of suitable gradients (and the boundedness of
certain random variables), we require that the parameterized class of
stochastic matrices satisfies the following additional assumption.
\begin{assumption}
\label{ass:pbound}
The derivatives,
$$
\nabla P(\theta) := \[\frac{\partial p_{ij}(\theta)}{\partial
        \theta_k}\]_{i,j = 1\dots n; k=1\dots K} 
$$
exist for all $\theta \in \R^K$.
The ratios 
$$
\[\frac{\left|\frac{\partial p_{ij}(\theta)}{\partial
\theta_k}\right|}{p_{ij}(\theta)}\]_{i,j = 1\dots n; k=1\dots K} 
$$
are uniformly bounded by $B < \infty$ for all $\theta\in \R^K$.
\end{assumption}
The second part of this assumption allows zero-probability transitions
$p_{ij}(\theta) = 0$ only if $\nabla p_{ij}(\theta)$ is also
zero, in which case we set $0/0 = 0$. One example is if $i\rightarrow
j$ is a forbidden transition, so that $p_{ij}(\theta) = 0$ for all
$\theta\in\R^K$. Another example satisfying the assumption is
$$
p_{ij}(\theta) = \frac{e^{\theta_{ij}}}{\sum_{j=1}^n e^{\theta_{ij}}}, 
$$
where $\theta = \[\theta_{11}, \dots, \theta_{1n}, \dots, \theta_{nn}\]
\in \R^{n^2}$ are the parameters of $P(\theta)$, for then 
\begin{align*}
\frac{\partial p_{ij}(\theta)/\partial \theta_{ij}}{p_{ij}(\theta)} &=
1 - p_{ij}(\theta),\quad\text{and}\\
\frac{\partial p_{ij}(\theta)/\partial \theta_{kl}}{p_{ij}(\theta)} &=
-p_{kl}(\theta).
\end{align*}

Assuming for the moment that $\nabla\pi(\theta)$ exists (this will
be justified shortly), then, suppressing $\theta$ dependencies,
\begin{equation}\label{eq:gradeta1}
\nabla \eta = \nabla \pi' r,
\end{equation}
since the reward $r$ does not depend on $\theta$. Note that our convention for
$\nabla$ in this paper is that it takes precedence over all other
operations, so $\nabla g(\theta) f(\theta) = \[\nabla g(\theta)\]
f(\theta)$. 
Equations like \eqref{eq:gradeta1} should be regarded as shorthand notation for 
$K$ equations of the form 
$$
\frac{\partial \eta(\theta)}{\partial \theta_k} = \[\frac{\partial
        \pi(\theta, 1)}{\partial \theta_k}, \dots, \frac{\partial
        \pi(\theta, n)}{\partial \theta_k}\] \[r(1), \dots, r(n)\]' 
$$
where $k=1,\dots, K$. To compute $\nabla \pi$, first differentiate the balance
equations~\eqref{eq:balance} to obtain
  $$
    \nabla\pi' P + \pi' \nabla P = \nabla\pi',
  $$
and hence
  \begin{equation}\label{eq:gradpi1}
    \nabla\pi' (I - P) = \pi' \nabla P.
  \end{equation}
The system of equations defined by~\eqref{eq:gradpi1} is under-constrained because $I
- P$ is not invertible (the balance equations show that $I - P$ has a left
eigenvector with zero eigenvalue). However, let $e$ denote the
$n$-dimensional column vector consisting of all $1$s, so that
$e\pi'$ is the $n\times n$ matrix with the stationary
distribution $\pi'$ in each row. Since $\nabla \pi' e =
\nabla (\pi' e) = \nabla (1) = 0$, we can rewrite \eqref{eq:gradpi1} as
$$
\nabla\pi'\[I -(P - e\pi')\] = \pi'\nabla P.
$$

To see that the inverse $\[I -(P - e\pi')\]^{-1}$ exists, let $A$ be
any matrix satisfying $\lim_{t\to\infty} A^t=0$. Then we can write
  \begin{align*}
    \lim_{T\to\infty} \[(I-A)\sum_{t=0}^T A^t \]
      & = \lim_{T\to\infty} \[\sum_{t=0}^T A^t - \sum_{t=1}^{T+1} A^t \] \\
      & = I - \lim_{T\to\infty} A^{T+1} \\
      & = I.
  \end{align*}
Thus,
  $$
    (I-A)^{-1}=\sum_{t=0}^\infty A^t.
  $$
It is easy to prove by induction that $\[P - e\pi'\]^t = P^t -
e\pi'$ which converges to $0$ as $t\rightarrow \infty$ by Assumption
\ref{ass:station}. So $\[I -(P - e\pi')\]^{-1}$ exists and is equal to
$\sum_{t=0}^\infty \[P^t - e\pi'\]$. Hence, we can write
  \begin{equation}\label{eq:gradpi}
    \nabla \pi' = \pi'\nabla P\[I - P + e \pi'\]^{-1},
  \end{equation}
  and so\footnote{The argument leading to \eqref{eq:gradpi} coupled
    with the fact that $\pi(\theta)$ is the unique solution to
    \eqref{eq:balance} can be used to justify the existence of
    $\nabla\pi$. Specifically, we can run through the same steps
    computing the value of $\pi(\theta + \delta)$ for small $\delta$
    and show that the expression \eqref{eq:gradpi} for $\nabla\pi$ is
    the unique matrix satisfying $\pi(\theta + \delta) = \pi(\theta) +
    \delta\nabla\pi(\theta) + O(\|\delta\|^2)$.}
\begin{equation}
\label{eq:gradeta}
\nabla \eta = \pi'\nabla P\[I - P + e \pi'\]^{-1} r.
\end{equation}
For \mdps\ with a sufficiently small number of states, \eqref{eq:gradeta}
could be solved exactly to yield the precise gradient direction.  However,
in general, if the state space is small enough that an exact solution of
\eqref{eq:gradeta} is possible, then it will be small enough to derive the
optimal policy using policy iteration and table-lookup, and there would
be no point in pursuing a gradient based approach in the first
place\footnote{Equation \eqref{eq:gradeta} may still be useful for \pomdps,
since in that case there is no tractable dynamic programming algorithm.}.

Thus, for problems of practical interest, \eqref{eq:gradeta} will be
intractable and we will need to find some other way of computing the
gradient. One approximate technique for doing this is presented in the
next section.

\section{Approximating the Gradient in Parameterized Markov Chains}
\label{sec:approx}

In this section, we show that the gradient can be split into two
components, one of which becomes negligible as a discount factor
$\beta$ approaches $1$. 

For all $\beta\in[0,1)$, let $J_\beta(\theta) = \[J_\beta(\theta,1), \dots,
J_\beta(\theta,n)\]$ denote the vector of expected discounted rewards from
each state $i$:
\begin{equation}
\label{eq:expdisc}
J_\beta(\theta,i) := \E_\theta\[\left.
  \sum_{t=0}^\infty \beta^t r(X_t) \right| X_0 = i\]. 
\end{equation}
Where the $\theta$ dependence is obvious, we just write $J_\beta$. 
\begin{proposition}
\label{theorem:factor}
For all $\theta\in\R^K$ and $\beta \in [0,1)$, 
\begin{equation}
\label{eq:factor}
\nabla\eta = (1-\beta)\nabla\pi'J_\beta +
\beta\pi' \nabla P J_\beta.
\end{equation}
\end{proposition}
\begin{proof}
Observe that $J_\beta$ satisfies
the {\em Bellman} equations:
\begin{equation}
\label{eq:bell}
J_\beta = r + \beta P J_\beta.
\end{equation}
\cite{bertsekas95}. Hence, 
\begin{align*}
\nabla \eta
        &= \nabla \pi' r \\
        &= \nabla \pi' \[J_\beta - \beta P J_\beta\]\\
        &= \nabla \pi' J_\beta - \beta \nabla\pi' J_\beta +
        \beta \pi'\nabla P J_\beta  &\text{\hfill by \eqref{eq:gradpi1}}\\
        &= (1 - \beta) \nabla\pi' J_\beta + \beta \pi' \nabla P
        J_\beta.
\end{align*}
\end{proof}
We shall see in the next section that the second term
in~\eqref{eq:factor} can be estimated from a single sample path of the
Markov chain.  In fact, Theorem~1 in \cite{kimura97} shows that the
gradient estimates of the algorithm presented in that paper converge
to $(1-\beta)\pi'\nabla J_\beta$.  By the Bellman
equations~\eqref{eq:bell}, this is equal to $(1-\beta)\beta(\pi'\nabla
PJ_\beta + \pi'\nabla J_\beta)$, which implies $(1-\beta)\pi'\nabla
J_\beta = \beta\pi'\nabla P J_\beta$.  Thus the algorithm of
\citeA{kimura97} also estimates the second term in the expression for
$\nabla \eta(\theta)$ given by~\eqref{eq:factor}.  It is important to
note that $\pi'\nabla J_\beta \neq \nabla \[\pi' J_\beta\]$---the two
quantities disagree by the first term in \eqref{eq:factor}. This
arises because the the stationary distribution itself depends on the
parameters. Hence, the algorithm of \citeA{kimura97} does not estimate
the gradient of the expected discounted reward.
In fact, the expected discounted reward is
simply $1/(1-\beta)$ times the average reward $\eta(\theta)$
\shortcite[Fact 7]{singh94a}, so the gradient of the expected discounted
reward is proportional to the gradient of the average reward.

The following theorem shows that the first term in~\eqref{eq:factor}
becomes negligible as $\beta$ approaches $1$.  Notice that this is not
immediate from Proposition~\ref{theorem:factor}, since $J_\beta$ can become
arbitrarily large in the limit $\beta\to 1$.

\begin{theorem}
\label{theorem:approx}
For all $\theta\in\R^K$,
\begin{equation}
\label{eq:approx}
\nabla \eta = \lim_{\beta\to 1}\nb \eta,
\end{equation}
where 
\begin{equation}
\label{eq:approxgrad}
\nb \eta := \pi'\nabla P J_\beta.
\end{equation}
\end{theorem}
\begin{proof}
Recalling equation \eqref{eq:gradeta} and the discussion preceeding
it, we have\footnote{Since $e\pi' r = e\eta$, \eqref{eq:grad2}
motivates a different kind of algorithm for estimating $\nabla\eta$
based on {\em differential rewards} \cite{marbach98}.}
\begin{equation}
\label{eq:grad2}
\nabla\eta = \pi'\nabla P \sum_{t=0}^\infty \[P^t - e\pi'\]r. 
\end{equation}
But $\nabla P e = \nabla(Pe) = \nabla(1) = 0$ since $P$ is a
stochastic matrix, so \eqref{eq:grad2} can be rewritten as 
\begin{equation}
\label{eq:grad3}
\nabla\eta= \pi'\[\sum_{t=0}^\infty \nabla P P^t\] r. 
\end{equation}
Now let $\beta\in [0,1]$ be a discount factor and consider the expression
\begin{equation}
\label{eq:grad4}
f(\beta) := \pi'\[\sum_{t=0}^\infty \nabla P (\beta P)^t\] r
\end{equation}
Clearly $\nabla\eta = \lim_{\beta\rightarrow 1} f(\beta)$. To complete
the proof we just need to show that $f(\beta) = \nb\eta$. 

Since $(\beta P)^t = \beta^tP^t \rightarrow \beta^t e\pi' \rightarrow
0$, we can invoke the observation before~\eqref{eq:gradpi} to write 
$$
\sum_{t=0}^\infty (\beta P)^t = \[I - \beta P \]^{-1}. 
$$
In particular, $\sum_{t=0}^\infty (\beta P)^t$ converges, so we 
can take $\nabla P$ back out of the sum in the right-hand-side of \eqref{eq:grad4} and 
write\footnote{We cannot back $\nabla P$ out of the sum in the
right-hand-side of \eqref{eq:grad3} because $\sum_{t=0}^\infty P^t$
diverges ($P^t \rightarrow e\pi'$). 
The reason $\sum_{t=0}^\infty\nabla PP^t$ converges is that $P^t$ becomes orthogonal to $\nabla P$
in the limit of large $t$. Thus, we can view $\sum_{t=0}^\infty P^t$
as a sum of two orthogonal components: an infinite one in the
direction $e$ and a finite one in the direction $e^\perp$. It is the
finite component that we need to estimate. Approximating 
$\sum_{t=0}^\infty P^t$ with $\sum_{t=0}^\infty (\beta P)^t$ is a way
of rendering the $e$-component finite while hopefully not
altering the $e^\perp$-component too much. There should be other
substitutions that lead to better approximations (in this context, see
the final paragraph in Section \ref{subsec:history}).}
\begin{equation}
\label{eq:grad5}
f(\beta) = \pi'\nabla P\[\sum_{t=0}^\infty \beta^t P^t\] r. 
\end{equation}
But $\[\sum_{t=0}^\infty \beta^t P^t\] r = J_\beta$. Thus $f(\beta) =
\pi'\nabla P J_\beta = \nb\eta$. 
\end{proof}

%

Theorem~\ref{theorem:approx} shows that $\nb\eta$ is a good approximation
to the gradient as $\beta$ approaches $1$, but it turns out that values
of $\beta$ very close to $1$ lead to large variance in the estimates of
$\nb\eta$ that we describe in the next section.  However, the following
theorem shows that $1-\beta$ need not be too small, provided the
transition probability matrix $P(\theta)$ has distinct eigenvalues, and
the Markov chain has a short {\em mixing time}. From any initial state,
the distribution over states of a Markov chain converges to the stationary
distribution, provided the assumption (Assumption~\ref{ass:station}) about
the existence and uniqueness of the stationary distribution is satisfied
\cite<see, for example,>[Theorem~15.8.1, p.~552]{lancaster85}. The
spectral resolution theorem~\cite[Theorem~9.5.1, p.~314]{lancaster85}
implies that the distribution converges to stationarity at an exponential
rate, and the time constant in this convergence rate (the mixing time)
depends on the eigenvalues of the transition probability matrix.
The existence of a unique stationary distribution implies that the
largest magnitude eigenvalue is $1$ and has multiplicity $1$, and the
corresponding left eigenvector is the stationary distribution. We sort
the eigenvalues $\lambda_i$ in decreasing order of magnitude, so that
$1=\lambda_1>|\lambda_2|>\cdots>|\lambda_s|$ for some $2\le s\le n$. It
turns out that $|\lambda_2|$ determines the mixing time of the chain.

The following theorem shows that if $1-\beta$ is small compared to
$1-|\lambda_2|$, the gradient approximation described above is accurate.
Since we will be using the estimate as a direction in which to update the
parameters, the theorem compares the {\em directions} of the gradient and
its estimate.  In this theorem, $\kappa_2(A)$ denotes the {\em spectral
condition number} of a nonsingular matrix $A$, which is defined as the
product of the {\em spectral norms} of the matrices $A$ and $A^{-1}$,
$$
\kappa_2(A) = \|A\|_2 \|A^{-1}\|_2,
$$
where 
$$
\|A\|_2 = \max_{x:\|x\|=1}\|Ax\|,
$$
and $\|x\|$ denotes the Euclidean norm of the vector $x$.

\begin{theorem}
\label{theorem:mix}
Suppose that the transition probability matrix $P(\theta)$
satisfies Assumption~\ref{ass:station} with stationary distribution
$\pi'=(\pi_1,\ldots,\pi_n)$, and has $n$ distinct eigenvalues.
Let $S=(x_1x_2\cdots x_n)$ be the matrix of right eigenvectors of $P$
corresponding, in order, to the eigenvalues
$1=\lambda_1>|\lambda_2|\ge\cdots\ge|\lambda_n|$.
Then the normalized inner product between $\nabla \eta$ and
$\beta \nb\eta$ satisfies
\begin{equation}\label{eqn:direrr}
1 - \frac{\nabla\eta\cdot\beta\nb\eta}{\|\nabla\eta\|^2}
\le \kappa_2\left(\Pi^{1/2}S\right)
\frac{\|\nabla(\sqrt\pi_1,\ldots,\sqrt\pi_n)\|}{\|\nabla\eta\|}
\sqrt{r'\Pi r} \frac{1-\beta}{1-\beta|\lambda_2|},
\end{equation}
where $\Pi=\diag(\pi_1,\ldots,\pi_n)$.
\end{theorem}

Notice that $r'\Pi r$ is the expectation under the stationary
distribution of $r(X)^2$.

As well as the mixing time (via $|\lambda_2|$), the bound in the
theorem depends on another parameter of the Markov chain: the spectral
condition number of $\Pi^{1/2}S$.  If the Markov chain is reversible
(which implies that the eigenvectors $x_1, \dots, x_n$ are
orthogonal), this is equal to the ratio of the maximum to the minimum
probability of states under the stationary distribution. However, the
eigenvectors do not need to be nearly orthogonal.  In fact, the
condition that the transition probability matrix have $n$ distinct
eigenvalues is not necessary; without it, the condition number is
replaced by a more complicated expression involving spectral norms of
matrices of the form $(P-\lambda_iI)$.

\begin{proof}
The existence of $n$ distinct eigenvalues implies that $P$ can be
expressed as $S\Lambda S^{-1}$, where
$\Lambda=\diag(\lambda_1,\ldots,\lambda_n)$ \cite[Theorem~4.10.2,
p~153]{lancaster85}.  It follows that for any polynomial $f$, we can
write $f(P)=Sf(\Lambda)S^{-1}$.

Now, Proposition~\ref{theorem:factor} shows that
$\nabla\eta-\beta\nb\eta=\nabla\pi'(1-\beta)J_\beta$.
But
\begin{align*}
(1-\beta)J_\beta
 &=  (1-\beta)\left(r + \beta P r + \beta^2 P^2 r + \cdots\right) \\
 &=  (1-\beta)\left(I + \beta P + \beta^2 P^2 + \cdots\right)r \\
 &=  (1-\beta)S\left(\sum_{t=0}^\infty \beta^t \Lambda^t\right)
   S^{-1} r \\
 &= (1-\beta)\sum_{j=1}^n x_j y_j'\left(\sum_{t=0}^\infty
   (\beta \lambda_j)^t\right) r,
\end{align*}
where $S^{-1} = ( y_1, \ldots, y_n)'$.

It is easy to verify that $y_i$ is the left eigenvector corresponding
to $\lambda_i$, and that we can choose $y_1=\pi$ and $x_1=e$. Thus we
can write
\begin{align*}
(1-\beta)J_\beta
 &= (1 - \beta) e\pi' r + \sum_{j=2}^n x_j y_j'\left(\sum_{t=0}^\infty
   (1-\beta)(\beta\lambda_j)^t\right) r \\
 &= (1-\beta) e\eta + \sum_{j=2}^n x_j y_j'
   \left(\frac{1-\beta}{1-\beta\lambda_j}\right) r \\
 &= (1-\beta) e\eta + S M S^{-1} r,
\end{align*}
where
$$
M = \diag\left(0,\frac{1-\beta}{1-\beta\lambda_2}, \ldots,
  \frac{1-\beta}{1-\beta\lambda_n}\right).
$$
It follows from this and Proposition~\ref{theorem:factor} that
\begin{align*}
1 - \frac{\nabla\eta\cdot\beta\nb\eta}{\|\nabla\eta\|^2}
&=  1 -
  \frac{\nabla\eta\cdot\left(\nabla\eta -
    \nabla\pi'(1-\beta) J_\beta\right)}{\|\nabla\eta\|^2} \\
&=  \frac{\nabla\eta\cdot \nabla\pi'(1-\beta) J_\beta}
  {\|\nabla\eta\|^2} \\
&=  \frac{\nabla\eta\cdot \nabla\pi'\left((1-\beta)e\eta+SMS^{-1}r\right)}
  {\|\nabla\eta\|^2} \\
&=  \frac{\nabla\eta\cdot \nabla\pi'SMS^{-1}r} {\|\nabla\eta\|^2} \\
&\le \frac{\left\|\nabla\pi'SMS^{-1}r\right\|}{\|\nabla\eta\|},
\end{align*}
by the Cauchy-Schwartz inequality.  Since $\nabla\pi' =
\nabla\left(\sqrt{\pi'}\right)\Pi^{1/2}$, we can apply
the Cauchy-Schwartz inequality again to obtain
\begin{equation}\label{eqn:blubsy}
1 - \frac{\nabla\eta\cdot\beta\nb\eta}{\|\nabla\eta\|^2}
\le \frac{\left\|\nabla\left(\sqrt{\pi'}\right)\right\|
  \left\|\Pi^{1/2} SMS^{-1}r\right\|}
  {\|\nabla\eta\|}.
\end{equation}
We use spectral norms to bound the second factor in the numerator.
It is clear from the definition that the spectral norm of a product of
nonsingular matrices satisfies $\|AB\|_2\le\|A\|_2\|B\|_2$, and
that the spectral norm of a diagonal matrix is given by
$\|\diag(d_1,\ldots,d_n)\|_2 = \max_i|d_i|$.  It follows that
\begin{align*}
\left\|\Pi^{1/2} SMS^{-1}r\right\|
&=
  \left\|\Pi^{1/2} SMS^{-1}\Pi^{-1/2}\Pi^{1/2}r\right\| \\
&\le
  \left\|\Pi^{1/2} S\right\|_2 \left\|S^{-1}\Pi^{-1/2}\right\|_2
  \left\|\Pi^{1/2}r\right\|\|M\|_2 \\
&\le 
  \kappa_2\left(\Pi^{1/2} S\right)\sqrt{r'\Pi r}
  \frac{1-\beta}{1-\beta|\lambda_2|}.
\end{align*}
Combining with Equation~\eqref{eqn:blubsy} proves~\eqref{eqn:direrr}.
\end{proof}

\section{Estimating the Gradient in Parameterized Markov Chains}
\label{section:comp}

Algorithm \ref{algorithm:gradmdp} introduces \mcg\ ({\bf M}arkov {\bf C}hain
{\bf G}radient), an algorithm
for estimating the approximate gradient $\nb\eta$ from a
single on-line sample path $X_0, X_1, \dots$ from the Markov chain
$M(\theta)$. 
\mcg\ requires only $2K$ reals to be stored, where $K$ is the
dimension of the parameter space: $K$ parameters for the eligibility
trace $z_t$, and $K$ parameters for the gradient estimate
$\Delta_t$. Note that after $T$ time steps $\Delta_T$ is the
average so far of $r(X_t) z_t$,
$$
\Delta_T = \frac1T\sum_{t=0}^{T-1} z_t r(X_t).
$$

\begin{algorithm}
\caption{The \mcg\ ({\bf M}arkov {\bf C}hain
{\bf G}radient) algorithm}
\label{algorithm:gradmdp}
\begin{algorithmic}[1]
\STATE {\bf Given: }
\begin{itemize}
\item Parameter $\theta\in\R^K$.
\item Parameterized class of stochastic matrices $\P =
\{P(\theta)\colon \theta\in\R^K\}$ satisfying Assumptions
\ref{ass:pbound} and \ref{ass:station}.
\item $\beta \in [0,1)$.
\item Arbitrary starting state $X_0$.
\item State sequence $X_0, X_1, \dots$ generated by $M(\theta)$
(i.e. the Markov chain with transition probabilities $P(\theta)$). 
\item Reward sequence $r(X_0), r(X_1), \dots$ satisfying
Assumption~\ref{ass:rbound}.
\end{itemize}
\STATE Set $z_0 = 0$ and $\Delta_0 = 0$ ($z_0, \Delta_0 \in\R^K$). 
\FOR{each state $X_{t+1}$ visited}
\STATE $z_{t+1} = \beta z_t +
  \dfrac{\nabla p_{X_t X_{t+1}}(\theta)}{p_{X_t X_{t+1}}(\theta)}$
\STATE $\Delta_{t+1} = \Delta_t + \frac1{t+1}\[r(X_{t+1}) z_{t+1} - \Delta_t\]$
\ENDFOR
\end{algorithmic}
\end{algorithm}

\begin{theorem}
\label{theorem:gradmdp}
Under Assumptions~\ref{ass:station},~\ref{ass:rbound}
and~\ref{ass:pbound}, the \mcg\ algorithm starting from any initial state
$X_0$ will generate a sequence $\Delta_0, \Delta_1, \dots, \Delta_t,
\dots$ satisfying
\begin{equation}
\label{converge}
\lim_{t\to\infty} \Delta_t = \nb\eta \quad\text{\rm{w.p.1}}.
\end{equation}
\end{theorem}

\begin{proof}
Let $\{X_t\} = \{X_0, X_1, \dots\}$ denote the random process
corresponding to $M(\theta)$. If $X_0\sim \pi$ then the entire process
is stationary. The proof can easily be generalized to arbitrary
initial distributions using the fact that under
Assumption~\ref{ass:station}, $\{X_t\}$ is asymptotically
stationary. When $\{X_t\}$ is stationary, we can write
\begin{align}
\pi'\nabla P J_\beta 
        &= \sum_{i,j} \pi(i) \nabla p_{ij}(\theta) J_\beta(j) \notag\\
        &= \sum_{i,j} \pi(i) p_{ij}(\theta)
                \frac{\nabla p_{ij}(\theta)}{p_{ij}(\theta)} J_\beta(j)
                \notag\\
        &= \sum_{i,j} \Pr(X_t=i) \Pr(X_{t+1}=j|X_t=i)
                \frac{\nabla p_{ij}(\theta)}{p_{ij}(\theta)}
                \Expect(J(t+1)|X_{t+1}=j),
\label{eq:poobum}
\end{align}
where the first probability is with respect to the stationary distribution
and $J(t+1)$ is the process
$$
J(t+1) = \sum_{s=t+1}^\infty \beta^{s-t-1} r(X_s).
$$
The fact that $\E(J(t+1)|X_{t+1}) = J_\beta(X_{t+1})$ for all
$X_{t+1}$ follows from the boundedness of the magnitudes of the
rewards (Assumption~\ref{ass:rbound}) and Lebesgue's dominated convergence
theorem. We can rewrite Equation~\eqref{eq:poobum} as
$$
\pi'\nabla P J_\beta 
        = \sum_{i,j} \E \[\chi_i(X_t) \chi_j(X_{t+1}) 
                \frac{\nabla p_{ij}(\theta)}{p_{ij}(\theta)}
                J(t+1)\],
$$
where $\chi_i(\cdot)$ denotes the indicator function for state $i$,
$$
\chi_i(X_t) := \begin{cases} 1 \quad \text{if $X_t = i$}, \\
                             0 \quad \text{otherwise},
                \end{cases}
$$
and the expectation is again with respect to the stationary
distribution. When $X_t$ is chosen according to the stationary
distribution, the process $\{X_t\}$ is ergodic. Since the
process $\{Z_t\}$ defined by
$$ 
Z_t := \chi_i(X_t) \chi_j(X_{t+1})
        \frac{\nabla p_{ij}(\theta)}{p_{ij}(\theta)} J(t+1)
$$
is obtained by taking a fixed function of $\{X_t\}$, $\{Z_t\}$ is
also stationary and ergodic \cite[Proposition 6.31]{brieman66}.
Since $\left|\frac{\nabla p_{ij}(\theta)}{p_{ij}(\theta)}\right|$ is
bounded by Assumption~\ref{ass:pbound}, from the ergodic theorem we have (almost surely):
\begin{align}
\pi'\nabla P J_\beta 
        &= \sum_{i,j} \lim_{T\rightarrow\infty}
                \frac1T\sum_{t=0}^{T-1} \chi_i(X_t) \chi_j(X_{t+1})
                \frac{\nabla p_{ij}(\theta)}{p_{ij}(\theta)}
                J(t+1) \notag \\
        &= \lim_{T\rightarrow\infty} \frac1T\sum_{t=0}^{T-1}
                \frac{\nabla p_{X_tX_{t+1}}(\theta)}{p_{X_tX_{t+1}}(\theta)}
                J(t+1) \notag\\
        &= \lim_{T\rightarrow\infty}\frac1T\sum_{t=0}^{T-1}
                \frac{\nabla p_{X_tX_{t+1}}(\theta)}{p_{X_tX_{t+1}}(\theta)}  
                \[ \sum_{s=t+1}^T \beta^{s-t-1} r(X_s)  + 
                \sum_{s=T+1}^{\infty}\beta^{s-t-1} r(X_s) \]. \label{eq:total}
\end{align}
Concentrating on the second term in the right-hand-side of \eqref{eq:total},
observe that:
\begin{align*}
\left|\frac1T\sum_{t=0}^{T-1}\frac{\nabla
                p_{X_tX_{t+1}}(\theta)}{p_{X_tX_{t+1}}(\theta)}  
                \right.&\left.\sum_{s=T+1}^{\infty}\beta^{s-t-1} r(X_s)\right| \\
        &\leq \frac1T\sum_{t=0}^{T-1}\left|\frac{\nabla
                p_{X_tX_{t+1}}(\theta)}{p_{X_tX_{t+1}}(\theta)}\right|  
                \sum_{s=T+1}^{\infty}\beta^{s-t-1}
                \left|r(X_s)\right| \\
        &\leq \frac{BR}{T}\sum_{t=0}^{T-1}\sum_{s=T+1}^{\infty}\beta^{s-t-1} \\
        &= \frac{BR}{T}\sum_{t=0}^{T-1}\frac{\beta^{T-t}}{1-\beta} \\
        &= \frac{BR\beta\(1 - \beta^T\)}{T\(1-\beta\)^2} \\
        &\rightarrow 0 \,\text{as $T\rightarrow\infty$}, 
\end{align*}
where $R$ and $B$ are the bounds on the magnitudes of the rewards
and $\frac{|\nabla p_{ij}|}{p_{ij}}$ from Assumptions~\ref{ass:rbound}
and~\ref{ass:pbound}.  Hence,
\begin{equation}
\label{eq:limit}
\pi'\nabla P J_\beta = \lim_{T\rightarrow\infty}\frac1T\sum_{t=0}^{T-1}
                \frac{\nabla p_{X_tX_{t+1}}(\theta)}{p_{X_tX_{t+1}}(\theta)}  
                \sum_{s=t+1}^T \beta^{s-t-1} r(X_s).
\end{equation}

Unrolling the equation for $\Delta_T$ in the  \mcg\ algorithm shows
it is equal to 
$$
\frac1T\sum_{t=0}^{T-1}\frac{\nabla p_{X_tX_{t+1}}(\theta)}{p_{X_tX_{t+1}}(\theta)} 
                \sum_{s=t+1}^T \beta^{s-t-1} r(i_s),
$$
hence $\Delta_T \rightarrow \pi'\nabla P J_\beta$ w.p.1 as
required. 
\end{proof}

\section{Estimating the Gradient in Partially Observable Markov
Decision Processes}
\label{section:comp1}

Algorithm \ref{algorithm:gradmdp} applies to any parameterized class of
stochastic matrices $P(\theta)$ for which we can compute the gradients
$\nabla p_{ij}(\theta)$. In this section we consider the special case of
$P(\theta)$ that arise from a parameterized class of randomized policies
controlling a partially observable Markov decision process (\pomdp).
The `partially observable' qualification means we assume that these
policies have access to an observation process that depends on the state,
but in general they may not see the state.

Specifically, assume that there are $N$ controls $\U = \{1, \dots, N\}$
and $M$ observations $\Y = \{1,\dots, M\}$.  Each $u\in \U$ determines
a stochastic matrix $P(u)$ which does not depend on the parameters
$\theta$.  For each state $i\in\S$, an observation $Y\in\Y$ is
generated independently according to a probability distribution $\nu(i)$
over observations in $\Y$.  We denote the probability of observation
$y$ by $\nu_y(i)$.  A {\em randomized policy} is simply a function
$\mu$ mapping observations $y\in\Y$ into probability distributions
over the controls $\U$.  That is, for each observation $y$, $\mu(y)$
is a distribution over the controls in $\U$.  Denote the probability
under $\mu$ of control $u$ given observation $y$ by $\mu_u(y)$.

To each randomized policy $\mu(\cdot)$ and observation distribution
$\nu(\cdot)$ there corresponds a Markov chain in which state transitions
are generated by first selecting an observation $y$ in state $i$
according to the distribution $\nu(i)$, then selecting a control $u$
according to the distribution $\mu(y)$, and then generating a transition
to state $j$ according to the probability $p_{ij}(u)$. To parameterize
these chains we parameterize the policies, so that $\mu$ now becomes
a function $\mu(\theta, y)$ of a set of parameters $\theta\in\R^K$ as
well as the observation $y$. The Markov chain corresponding to $\theta$
has state transition matrix $[p_{ij}(\theta)]$ given by
\begin{equation}
\label{eq:barf}
p_{ij}(\theta) = \E_{Y\sim \nu(i)} \E_{U\sim \mu(\theta, Y)} p_{ij}(U). 
\end{equation}
Equation \eqref{eq:barf} implies 
\begin{equation}
\label{eq:barf1}
\nabla p_{ij}(\theta) = \sum_{u,y} \nu_y(i) p_{ij}(u) \nabla \mu_u(\theta, y). 
\end{equation}

Algorithm \ref{algorithm:pgradmdp} introduces the \pomdpg\ algorithm
(for {\bf G}radient of a {\bf P}artially {\bf O}bservable {\bf M}arkov {\bf D}ecision {\bf
P}rocess), a modified form of Algorithm
\ref{algorithm:gradmdp} in which updates of $z_t$ are based on
$\mu_{U_t} (\theta, Y_t)$, rather than $p_{X_t X_{t+1}}(\theta)$.
Note that Algorithm~\ref{algorithm:pgradmdp} does not require
knowledge of the transition probability matrix $P$, nor of the
observation process $\nu$; it only requires knowledge of the
randomized policy $\mu$. \pomdpg\ is essentially the algorithm
proposed by \citeA{kimura97} without the reward baseline. 

The algorithm \pomdpg\ assumes that the policy $\mu$ is a function only of
the current observation. It is immediate that the same algorithm works for
any finite history of observations. In general, an optimal policy needs to
be a function of the entire observation history. \pomdpg\ can be extended
to apply to policies with internal state \cite{tr_belief_01}.

\begin{algorithm}
\caption{The \pomdpg\ algorithm.}
\label{algorithm:pgradmdp}
\begin{algorithmic}[1]
\STATE {\bf Given: }
\begin{itemize}
\item Parameterized class of randomized policies $\left\{\mu(\theta, \cdot):
\theta \in \R^K\right\}$ satisfying Assumption~\ref{ass:mubound}.
\item Partially observable Markov decision process which when controlled
by the randomized policies $\mu(\theta, \cdot)$ corresponds to a
parameterized class of Markov chains satisfying Assumption~\ref{ass:station}.
\item $\beta \in [0,1)$.
\item Arbitrary (unknown) starting state $X_0$.
\item Observation sequence $Y_0, Y_1, \dots$  generated by the \pomdp\
with controls $U_0, U_1, \dots$ generated randomly according
to $\mu(\theta, Y_t)$. 
\item Reward sequence $r(X_0),r(X_1),\dots$ satisfying
Assumption~\ref{ass:rbound}, where $X_0,X_1,\dots$ is the (hidden)
sequence of states of the Markov decision process.
\end{itemize}
\STATE Set $z_0 = 0$ and $\Delta_0 = 0$ ($z_0, \Delta_0 \in\R^K$). 
\FOR{each observation $Y_t$, control $U_t$, and subsequent reward
$r(X_{t+1})$}
\STATE $z_{t+1} = \beta z_t +
  \dfrac{\nabla \mu_{U_t}(\theta, Y_t)}{\mu_{U_t}(\theta, Y_t)}$
\STATE $\Delta_{t+1} = \Delta_t + \frac1{t+1}\[r(X_{t+1}) z_{t+1}
  - \Delta_t\]$
\ENDFOR
\end{algorithmic}
\end{algorithm}
For convergence of Algorithm \ref{algorithm:pgradmdp}
we need to replace Assumption \ref{ass:pbound} with a similar bound
on the gradient of $\mu$: 
\begin{assumption}
The derivatives,
$$
\frac{\partial \mu_u(\theta, y)}{\partial \theta_k}
$$
exist for all $u\in \U$, $y\in \Y$ and $\theta \in \R^K$. 
\label{ass:mubound}
The ratios 
$$
\[\frac{\left|\frac{\partial \mu_u(\theta, y)}{\partial
\theta_k}\right|}{\mu_u(\theta, y)}\]_{y = 1\dots M; u = 1\dots N; k=1\dots K} 
$$
are uniformly bounded by $B_\mu < \infty$ for all $\theta\in \R^K$.
\end{assumption}

\begin{theorem}
\label{theorem:pgradmdp}
Under Assumptions~\ref{ass:station},~\ref{ass:rbound} and~\ref{ass:mubound},
Algorithm \ref{algorithm:pgradmdp} starting from any initial state $X_0$ will
generate a sequence $\Delta_0, \Delta_1, \dots, \Delta_t, \dots$
satisfying
\begin{equation}
\label{pconverge}
\lim_{t\to\infty} \Delta_t = \nb\eta \quad\text{\rm{w.p.1}}.
\end{equation}
\end{theorem}
\begin{proof}
The proof follows the same lines as the proof of Theorem
\ref{theorem:gradmdp}. In this case,
\begin{align*}
\pi' \nabla P J_\beta
        &= \sum_{i,j} \pi(i) \nabla p_{ij}(\theta) J_\beta(j) \\
        &= \sum_{i,j,y,u} \pi(i) p_{ij}(u) \nu_y(i) \nabla \mu_u(\theta, y)
                J_\beta(j) \text{ from \eqref{eq:barf1}}\\
        &= \sum_{i,j,y,u} \pi(i) p_{ij}(u) \nu_y(i)
                \frac{\nabla \mu_u(\theta, y)}{\mu_u(\theta, y)}
                \mu_u(\theta, y) J_\beta(j), \\
        &= \sum_{i,j,y,u} \Expect Z'_t,
\end{align*}
where the expectation is with respect to the stationary distribution
of $\{X_t\}$, and the process $\{Z'_t\}$ is defined by
$$
Z'_t := \chi_i(X_t) \chi_j(X_{t+1}) \chi_u(U_t) \chi_y(Y_t)
\frac{\nabla \mu_u(\theta, y)}{\mu_u(\theta, y)} J(t+1), 
$$
where $U_t$ is the control process and $Y_t$ is the observation process.
The result follows from the same arguments used in the proof of
Theorem~\ref{theorem:gradmdp}.
\end{proof}

\subsection{Control dependent rewards}
\label{section:cdr}
There are many circumstances in which the rewards may themselves
depend on the controls $u$. For example, some controls may consume
more energy than others and so we may wish to add a penalty term to
the reward function in order to conserve energy. 
The simplest way to deal with this is to define for each
state $i$ the expected reward $\rbar(i)$ by
\begin{equation}
\label{rbar}
\rbar(i) = \E_{Y\sim \nu(i)} \E_{U\sim\mu(\theta, Y)} r(U, i),
\end{equation}
and then redefine $J_\beta$ in terms of $\rbar$: 
\begin{equation}
\label{Jbar} 
\Jbar_\beta(\theta, i) := \lim_{N\rightarrow\infty}\E_\theta
\[\left. \sum_{t=0}^N \beta^t\rbar(X_t) \right|
X_0 = i\],
\end{equation}
where the expectation is over all trajectories $X_0, X_1,
\dots$. The performance gradient then becomes 
$$
\nabla\eta = \nabla\pi' \rbar + \pi' \nabla \rbar,
$$ 
which can be approximated by 
$$
\nb \eta = \pi'\[\nabla P \Jbar_\beta + \nabla \rbar\],
$$
due to the fact that 
$\Jbar_\beta$ satisfies the Bellman equations
\eqref{eq:bell} with $\rbar$ replaced by $r$. 

For \pomdpg\ to take account of the dependence of $r$ on the controls, 
its fifth line should be replaced by
$$
\Delta_{t+1} = \Delta_t + \frac1{t+1} \[r(U_{t+1}, X_{t+1})
\(z_{t+1} + \frac{\nabla \mu_{U_{t+1}}(\theta,
Y_{t+1})}{\mu_{U_{t+1}}(\theta, Y_{t+1})}\) - \Delta_t\].
$$
It is straightforward to extend the proofs of Theorems
\ref{theorem:approx}, \ref{theorem:mix} and \ref{theorem:pgradmdp} to
this setting.

\subsection{Parameter dependent rewards}
\label{section:pdr}

It is possible to modify \pomdpg\ when the rewards themselves depend
directly on $\theta$.  In this case, the fifth line of \pomdpg\ is
replaced with
\begin{equation}
\label{eq:rtheta}
\Delta_{t+1} = \Delta_t + \frac1{t+1} \[r(\theta, X_{t+1}) z_{t+1} +
\nabla r(\theta, X_{t+1}) - \Delta_t\].
\end{equation}
Again, the convergence and approximation theorems will carry
through, provided $\nabla r(\theta,i)$ is uniformly
bounded. Parameter-dependent rewards have been considered by
\citeA{glynn90}, \citeA{marbach98}, and \citeA{baird98}. In particular,
\citeA{baird98} 
showed how suitable choices of $r(\theta, i)$ lead to a combination of
value and policy search, or ``\vaps''. For example, if $\Jt(\theta,
i)$ is an approximate value-function, then setting\footnote{The use of
rewards $r(\theta, X_t, X_{t-1})$ that depend on the current and
previous state does not substantially alter the analysis.}  
$$
r(\theta, X_{t},X_{t-1}) = -\frac12\[r(X_t) + \alpha\Jt(\theta, X_t) -
\Jt(\theta, X_{t-1})\]^2,
$$ 
where $r(X_t)$ is the usual reward and $\alpha\in[0,1)$ is a
discount factor, gives an update that seeks to minimize the expected
Bellman error 
\begin{equation}
\label{eq:berror}
\sum_{i=1}^n \pi(\theta, i) \[r(i) + \alpha\sum_{j=1}^n p_{ij}(\theta)
\Jt(\theta,j) - \Jt(\theta, i)\]^2. 
\end{equation}
This will have the effect of both minimizing the Bellman error in
$\Jt(\theta, i)$, and driving the system (via the policy) to states
with small Bellman error. The motivation behind such an approach can
be understood if one considers a $\Jt$ that has {\em zero} Bellman
error for all states. In that case a greedy policy derived from $\Jt$
will be optimal, and regardless of how the actual policy is
parameterized, the expectation of $z_t r(\theta,X_t, X_{t-1})$ will be
zero and so will be the gradient computed by \gpomdp. This kind of
update is known as an {\em actor-critic} algorithm \cite{BarSutAnd83},
with the policy playing the role of the actor, and the value function
playing the role of the critic. 

\subsection{Extensions to infinite state, observation, and control spaces}

The convergence proof for Algorithm \ref{algorithm:pgradmdp}
relied on finite state ($\S$), observation ($\Y$) and control
($\U$) spaces. However, it should be clear that with no modification
Algorithm~\ref{algorithm:pgradmdp} can be applied immediately to \pomdps\
with countably or uncountably infinite $\S$ and $\Y$, and countable
$\U$. All that changes is that $p_{ij}(u)$ becomes a {\em kernel}
$p(x,x',u)$ and $\nu(i)$ becomes a density on observations.  In addition,
with the appropriate interpretation of $\nabla \mu/\mu$, it can be applied
to uncountable $\U$. Specifically, if $\U$ is a subset of $\R^N$ then
$\mu(y,\theta)$ will be a probability {\em density} function on $\U$ with
$\mu_u(y, \theta)$ the density at $u$.  If $\U$ and $\Y$ are subsets of
Euclidean space (but $\S$ is a finite set), Theorem~\ref{theorem:pgradmdp}
can be extended to show that the estimates produced by this algorithm
converge almost surely to $\nb\eta$.  In fact, we can prove a more general
result that implies both this case of densities on subsets of $\R^N$
as well as the finite case of Theorem~\ref{theorem:pgradmdp}. We allow
$\U$ and $\Y$ to be general spaces satisfying the following topological
assumption.  (For definitions see, for example,~\cite{dudley89}.)

\begin{assumption}\label{ass:top}
The control space $\U$ has an associated topology that is separable,
Hausdorff, and first-countable. For the corresponding Borel
$\sigma$-algebra $\B$ generated by this topology, there is a
$\sigma$-finite measure $\lambda$ defined on the measurable space
$(\U,\B)$. We say that $\lambda$ is the {\em reference measure}
for $\U$.

Similarly, the observation space $\Y$ has a topology, Borel
$\sigma$-algebra, and reference measure satisfying the same
conditions.
\end{assumption}

In the case of Theorem~\ref{theorem:pgradmdp}, where $\U$ and $\Y$ are
finite, the associated reference measure is the counting measure. For
$\U=\R^N$ and $\Y=\R^M$, the reference measure is Lebesgue measure. We
assume that the distributions $\nu(i)$ and $\mu(\theta,y)$ are absolutely
continuous with respect to the reference measures, and the corresponding
Radon-Nikodym derivatives (probability masses in the finite case,
densities in the Euclidean case) satisfy the following assumption.

\begin{assumption}\label{ass:densmubound}
For every $y\in\Y$ and $\theta\in\R^K$, the probability measure
$\mu(\theta,y)$ is absolutely continuous with respect to the reference
measure for $\U$.
For every $i\in\S$, the probability measure $\nu(i)$ is absolutely
continuous with respect to the reference measure for $\Y$.

Let $\lambda$ be the reference measure for $\U$.
For all $u\in \U$, $y\in \Y$, $\theta \in \R^K$, and
$k\in\{1,\dots,K\}$, the derivatives
$$
\frac{\partial}{\partial \theta_k} \frac{d\mu(\theta, y)}{d\lambda}(u)
$$
exist and the ratios 
$$
\frac{\left|\frac{\partial}{\partial \theta_k}
        \frac{d\mu_u(\theta, y)}{d\lambda}(u)\right|}
     {\frac{d\mu_u(\theta, y)}{d\lambda}(u)}
$$
are bounded by $B_\mu < \infty$.
\end{assumption}

With these assumptions, we can replace $\mu$ in
Algorithm~\ref{algorithm:pgradmdp} with the Radon-Nikodym derivative
of $\mu$ with respect to the reference measure on $\U$. In this case,
we have the following convergence result. This generalizes
Theorem~\ref{theorem:pgradmdp}, and also applies to densities $\mu$ on
a Euclidean space $\U$.

\begin{theorem}\label{theorem:genpgradmdp}
Suppose the control space $\U$ and the observation space $\Y$
satisfy Assumption~\ref{ass:top} and let $\lambda$ be the reference
measure on the control space $\U$.  Consider
Algorithm~\ref{algorithm:pgradmdp} with
  $$
    \frac{\nabla\mu_{U_t}(\theta,Y_t)}{\mu_{U_t}(\theta,Y_t)}
  $$
replaced by
$$
\frac{\nabla\frac{d\mu(\theta,Y_t)}{d\lambda}(U_t)}
  {\frac{d\mu(\theta,Y_t)}{d\lambda}(U_t)}.
$$
Under Assumptions~\ref{ass:station},~\ref{ass:rbound}
and~\ref{ass:densmubound}, this algorithm, starting from any initial state
$X_0$ will generate a sequence $\Delta_0, \Delta_1, \dots, \Delta_t,
\dots$ satisfying
$$
\lim_{t\to\infty} \Delta_t = \nb\eta \quad\text{\rm{w.p.1}}.
$$
\end{theorem}
\begin{proof}
See Appendix \ref{section:genpgradmdpproof}
\end{proof}

\section{New Results} 
\label{section:newstuff}
Since the first version of this paper, we have extended \gpomdp\ to
several new settings, and also proved some new properties of the
algorithm. In this section we briefly outline these results. 

\subsection{Multiple Agents}

Instead of a single agent generating actions according to $\mu(\theta,
y)$, suppose we have multiple agents $i=1,\dots, n_a$, each with their
own parameter set $\theta^i$ and distinct observation of the
environment $y^i$, and that generate their own actions $u^i$ according
to a policy $\mu_{u^i}(\theta^i, y^i)$. If the agents all receive the
same reward signal $r(X_t)$ (they may be cooperating to solve the same
task, for example), then \gpomdp\ can be applied to the collective
\pomdp\ obtained by concatenating the observations, controls, and
parameters into single vectors $y =\[y^1,\dots,y^{n_a}\]$, $u = \[u^1,
\dots, u^{n_a}\]$, and $\theta =\[\theta^1, \dots,
\theta^{n_a}\]$ respectively. An easy calculation shows that the gradient estimate
$\Delta$ generated by \gpomdp\ in the collective case is precisely the
same as that obtained by applying \gpomdp\ to each agent
independently, and then concatenating the results. That is, $\Delta =
\[\Delta^1,\dots, \Delta^{n_a}\]$, where $\Delta^i$ is the estimate
produced by \gpomdp\ applied to agent $i$. This leads to an on-line
algorithm in which the agents adjust their parameters independently
and without any explicit communication, yet collectively the
adjustments are maximizing the global average reward.  For similar
observations in the context of \reinforce\ and \vaps, see
\citeA{peshkin00}. This algorithm gives a biologically plausible
synaptic weight-update rule when applied to networks of spiking
neurons in which the neurons are regarded as independent agents
\cite{tr_hebb_99}, and has shown some promise in a network routing
application \cite{tr_route_01}.

\subsection{Policies with internal states}

So far we have only considered purely {\em reactive} or {\em
memoryless} policies in which the chosen control is a function of only
the current observation. \gpomdp\ is easily extended to cover the case
of policies that depend on finite histories of observations $Y_t,
Y_{t-1}, \dots, Y_{t-k}$, but in general, for {\em optimal} control of
\pomdps, the policy must be a function of the {\em entire} observation
history. Fortunately, the observation history may be summarized
in the form of a {\em belief state} (the current distribution over
states), which is itself updated based only upon the current
observation, and knowledge of which is sufficient for optimal
behaviour \cite{sondik73,sondik78}. An extension of \gpomdp\
to policies with parameterized internal belief states is described by
\citeA{tr_belief_01}, similar in spirit to the extension of \vaps\ and
\reinforce\ described by \citeA{meuleau99}. 

\subsection{Higher-Order Derivatives}

\gpomdp\ can be generalized to compute estimates of second and higher-order
derivatives of the average reward (assuming they exist), still from a
single sample path of the underlying \pomdp. To see this for
second-order derivatives, observe that if $\eta(\theta) = \int
q(\theta,x) r(x)\,dx$ for some twice-differentiable
density $q(\theta, x)$ and performance measure $r(x)$, then 
$$
\nabla^2\eta(\theta) = 
\int r(x) \frac{\nabla^2 q(\theta, x)}{q(\theta,x)} q(\theta, x) \, dx
$$
where $\nabla^2$ denotes the matrix of second derivatives
(Hessian). It can be verified that 
\begin{equation}
\label{eq:qeq}
\frac{\nabla^2 q(\theta, x)}{q(\theta, x)} = \nabla^2\log q(\theta, x)
+ \[\nabla\log q(\theta, x)\]^2
\end{equation}
where the second term on the right-hand-side is the {\em outer
product} between $\nabla\log q(\theta, x)$ and itself (that is, the
matrix with entries $\partial/\partial\theta_i \log q(\theta,
x)\partial/\partial\theta_j \log q(\theta, x)$). Taking $x$ to be a
sequence of states $X_0, X_1, \dots, X_T$ between visits to a
recurrent state $i^*$ in a parameterized Markov chain (recall Section
\ref{section:regen}), we have $q(\theta, X) = \Pi_{t=0}^{T-1}
p_{X_t X_{t+1}}(\theta)$, which combined with \eqref{eq:qeq} yields
  $$
    \frac{\nabla^2 q(\theta, X)}{q(\theta, X)}
      = \sum_{t=0}^{T-1}
      \frac{\nabla^2p_{X_tX_{t+1}}(\theta)}{p_{X_tX_{t+1}}(\theta)} -
      \sum_{t=0}^{T-1}\[
      \frac{\nabla p_{X_tX_{t+1}}(\theta)}{p_{X_tX_{t+1}}(\theta)}\]^2 
      + \[\sum_{t=0}^{T-1}
      \frac{\nabla p_{X_tX_{t+1}}(\theta)}{p_{X_tX_{t+1}}(\theta)}\]^2
$$
(the squared terms in this expression are also outer products). From
this expression we can derive a \pomdpg-like algorithm for computing a
biased estimate of the Hessian $\nabla^2\eta(\theta)$, which involves
maintaining---in addition to the usual eligibility trace $z_{t}$---a
second {\em matrix} trace updated as follows: 
$$
Z_{t+1} = \beta Z_t + \frac{\nabla^2
p_{X_tX_{t+1}}(\theta)}{p_{X_tX_{t+1}}(\theta)} - \[\frac{\nabla
p_{X_tX_{t+1}}(\theta)}{p_{X_tX_{t+1}}(\theta)}\]^2. 
$$
After $T$ time steps the algorithm returns the average so far of
$r(X_t) \[Z_t + z_t^2\]$ where the second term is again an
outer product. Computation of higher-order derivatives could be used in
second-order gradient methods for optimization of policy parameters.

\subsection{Bias and Variance Bounds}

Theorem \ref{theorem:mix} provides a bound on the {\em bias} of
$\nb\eta(\theta)$ relative to $\nabla\eta(\theta)$ that applies when
the underlying Markov chain has distinct eigenvalues. We have extended
this result to arbitrary Markov chains \cite{jcss_01}.  However, the
extra generality comes at a price, since the latter bound involves the
number of states in the chain, whereas Theorem \ref{theorem:mix} does
not. The same paper also supplies a proof that the variance of \gpomdp\
scales as $1 / (1 - \beta)^2$, providing a formal justification for the
interpretation of $\beta$ in terms of bias/variance trade-off.

\section{Conclusion}
\label{section:conc}

We have presented a general algorithm (\mcg) for computing arbitrarily
accurate approximations to the gradient of the average reward in a
parameterized Markov chain. When the chain's transition matrix has
distinct eigenvalues, the accuracy of the approximation was shown to
be controlled by the size of the subdominant eigenvalue
$|\lambda_2|$. We showed how the algorithm could be modified to
apply to partially observable Markov decision processes controlled by
parameterized stochastic policies, with both discrete and continuous
control, observation and state spaces (\pomdpg). For the finite state case, we
proved convergence with probability 1 of both algorithms.

We briefly described extensions to multi-agent problems, policies with
internal state, estimating higher-order derivatives, generalizations
of the bias result to chains with non-distinct eigenvalues, and a new
variance result. There are many avenues for further
research. Continuous time results should follow as extensions of the
results presented here.  The \mcg\ and \pomdpg\ algorithms can be
applied to countably or uncountably infinite state spaces; convergence
results are also needed in these cases.

In the companion paper \cite{jair_01b}, we present experimental results
showing rapid convergence of the estimates generated by \pomdpg\ to
the true gradient $\nabla\eta$. We give on-line variants of the
algorithms of the present paper, and also variants of gradient ascent
that make use of the estimates of $\nb\eta$. We present
experimental results showing the effectiveness of these algorithms in
a variety of problems, including a three-state MDP, a nonlinear
physical control problem, and a call-admission problem.

\subsection*{Acknowledgements}
This work was supported by the Australian Research Council, and
benefited from the comments of several anonymous referees. Most of
this research was performed while the authors were with the Research
School of Information Sciences and Engineering, Australian National
University.

\appendix

\section{A Simple Example of Policy Degradation in Value-Function Learning}
\label{section:badshit}
Approximate value-function approaches to reinforcement work by
minimizing some form of error between the approximate value function
and the true value function. It has long been known that this may not
necessarily lead to improved policy performance from the new value
function. We include this appendix because it illustrates that this
phenomenon can occur in the simplest possible system, a two-state
\mdp, and also provides some geometric intuition for why the
phenomenon arises. 

\begin{figure}
\begin{center}
\includegraphics[scale=0.6]{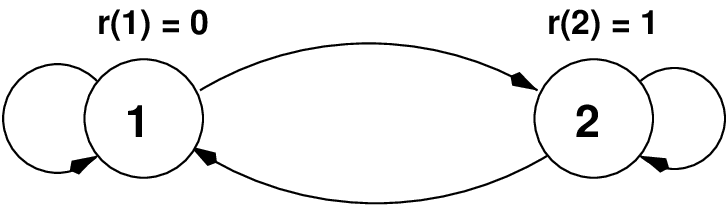}
\caption{Two-state Markov Decsision Process\label{figure:2state}} 
\end{center}
\end{figure}

Consider the two-state Markov decision process (MDP) in Figure
\ref{figure:2state}. There are two controls $u_1, u_2$ with
corresponding transition probability matrices $$ P(u_1) =
\begin{bmatrix} \frac13 & \frac23 \\ \frac13 & \frac23
\end{bmatrix},
\quad P(u_2) = \begin{bmatrix}
        \frac23 & \frac13 \\
        \frac23 & \frac13 
        \end{bmatrix},
$$
so that $u_1$ always takes the system to state $2$ with probability
$2/3$, regardless of the starting state (and therefore to state $1$
with probability $1/3$), and $u_2$ does the opposite. Since state $2$
has a reward of $1$, while state $1$ has a reward of $0$, the optimal
policy is to always select action $u_1$. Under this policy the
stationary distribution on states is $\[\pi_1,\pi_2\] = \[1/3,2/3\]$, while the
infinite-horizon discounted value of each state $i=1,2$ with discount
value $\alpha\in[0,1)$ is
$$ 
J_\alpha(i) = \E\(\left.\sum_{t=0}^\infty\alpha^tr(X_t)\right|X_0 = i\),
$$ 
where the expectation is over all state sequences $X_0, X_1, X_2,
\dots$ with state transitions generated according to $P(u_1)$. 
Solving {\em Bellman's equations:} $J_\alpha = r
+ \alpha P(u_1) J_\alpha$, where $J_\alpha = \[J_\alpha(1),
J_\alpha(2)\]'$ and $r = \[r(1), r(2)\]'$ yields $J_\alpha(1) =
\frac{2\alpha}{3(1-\alpha)}$ and $J_\alpha(2) = 1 +
\frac{2\alpha}{3(1-\alpha)}$. 

Now, suppose we are trying to learn an approximate value function
$\Jt$ for this \mdp, \ie, $\Jt(i) = w \phi(i)$ for each state $i=1,2$
and some scalar feature $\phi$ ($\phi$ must have dimensionality $1$ to
ensure that $\Jt$ really is {\em approximate}). Here $w\in\R$ is the
parameter being learnt. For the greedy policy obtained from $\Jt$ to be
optimal, $\Jt$ must value state $2$ above state $1$. For the
purposes of this illustration choose $\phi(1) = 2, \phi(2) = 1$, so
that for $\Jt(2) > \Jt(1)$, $w$ must be negative.

{\em Temporal Difference} learning (or $\TD(\lambda)$) is one of the
most popular techniques for training approximate value functions
\cite{sutton98}. It has been shown that for linear functions, $\TD(1)$
converges to a parameter $w^*$ minimizing the expected squared loss under 
the stationary distribution \cite{tsitsikilis97}:
\begin{equation}
\label{eq:loss}
w^* = \argmin_w \sum_{i=1}^2\pi_i \[w\phi(i) - J_\alpha(i)\]^2.
\end{equation}
Substituting the previous expressions for $\pi_1,\pi_2,\phi$ and $J_\alpha$
under the optimal policy and solving for $w^*$, yields $w^* = \frac{3 +
\alpha}{9(1 - \alpha)}$. Hence $w^* > 0$ for all values of
$\alpha\in[0,1)$, which is the wrong sign. So we have a situation
where the optimal policy is implementable as a greedy policy based on
an approximate value function in the class (just choose any $w<0$),
yet $\TD(1)$ observing the optimal policy will converge to a value
function whose corresponding greedy policy implements the suboptimal
policy.

\begin{figure}
\begin{center}
\includegraphics[scale=0.8]{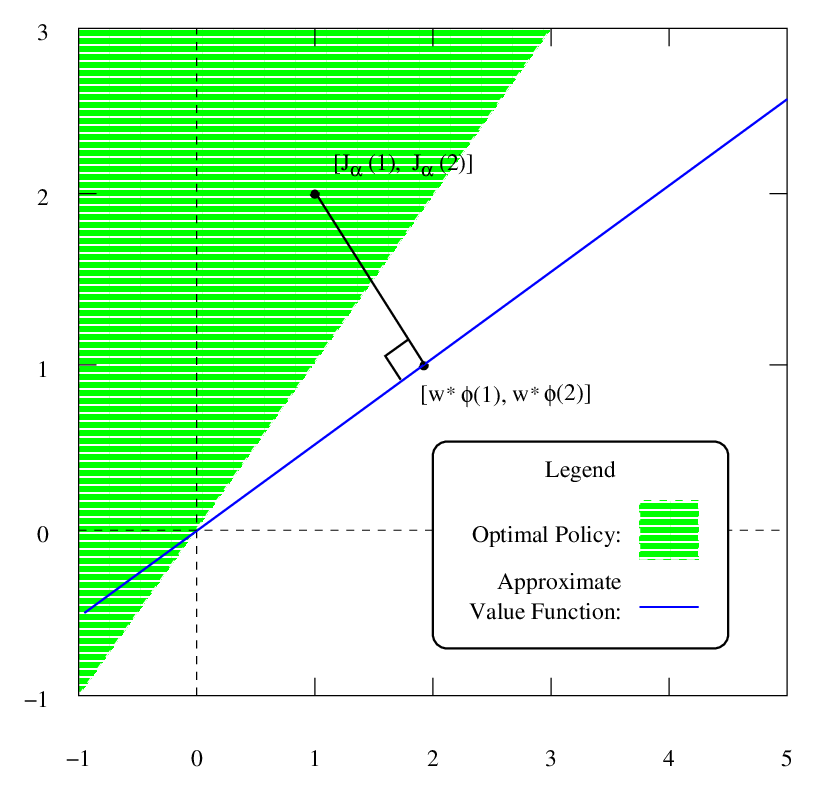}
\caption{Plot of value-function space for the two-state system. Note
that the scale of each axis has been weighted by the square root
of the stationary probability of the corresponding state under the
optimal policy. The solution found by TD(1) is simply the projection
of the true value function onto the set of approximate value
functions.
\label{figure:geometry}}
\end{center}
\end{figure}

A geometrical illustration of why this occurs is shown in Figure
\ref{figure:geometry}. In this figure, points on the graph represent
the values of the states. The scales of the state~1 and state~2 axes
are weighted by $\sqrt{\pi(1)}$ and $\sqrt{\pi(2)}$ respectively. In
this way, the squared euclidean distance on the graph between two
points $J$ and $\tilde J$ corresponds to the
expectation under the stationary distribution of the squared difference
between values:
  $$
    \left\| \[ \sqrt{\pi(1)} J(1), \sqrt{\pi(2)} J(2)\] - \[ \sqrt{\pi(1)}
    \tilde J(1), \sqrt{\pi(2)} \tilde J(2)\] \right\|^2 = \Expect_\pi \(
    J(X) - \tilde J(X) \)^2.
  $$
For any value function in the shaded region, the corresponding greedy
policy is optimal, since those value functions rank state~2 above
state~1. The bold line represents the set of all realizable approximate
value functions $(w\phi(1),w\phi(2))$. The solution to \eqref{eq:loss}
is then the approximate value function found by projecting the point
corresponding to the true value function $[(J_\alpha(1), J_\alpha(2)]$ onto
this line.  This is illustrated in the figure for $\alpha = 3/5$. The
projection is suboptimal because weighted mean-squared distance in
value-function space does not take account of the policy boundary.

\section{Proof of Theorem \ref{theorem:genpgradmdp}}
\label{section:genpgradmdpproof}
The proof needs the following topological lemma.  For definitions see,
for example,~\cite[pp.~24--25]{dudley89}.

\begin{lemma}\label{lem:topheavy}
Let $(X,\T)$ be a topological space that is Hausdorff, separable,
and first-countable. Let $\B$ be the Borel $\sigma$-algebra
generated by $\T$. Then the measurable space $(X,\B)$ has a
sequence $\S_1,\S_2,\ldots \subseteq\B$ of sets that
satisfies the following conditions:
\begin{enumerate}
\item
Each $\S_i$ is a partition of $X$ (that is, $X=\bigcup\{S\colon S\in\S_i\}$ and
any two distinct elements of $\S_i$ have empty intersection).
\item
For all $x\in X$, $\{x\}\in\B$ and
$$
\bigcap_{i=1}^\infty \{S\in\S_i: x\in S\} = \{x\}.
$$
\end{enumerate}
\end{lemma}

\begin{proof}
Since $X$ is separable, it has a countable dense subset
$S=\{x_1,x_2,\ldots\}$. Since $X$ is first-countable,
each of these $x_i$ has a countable neighbourhood base, $N_i$. Now,
construct the partitions $\S_i$ using the countable set
$N=\bigcup_{i=1}^\infty N_i$ as follows. Let $\S_0=X$ and, for
$i=1,2,\ldots$, define
$$
\S_i = \left\{S\cap N_i:S\in\S_{i-1}\right\} \cup
\left\{S\cap (X-N_i):S\in\S_{i-1}\right\}.
$$
Clearly, each $\S_i$ is a measurable partition of $X$.
Since $X$ is Hausdorff, for each pair $x,x'$ of distinct points from
$X$, there is a pair of disjoint open sets $A$ and $A'$ such that
$x\in A$ and $x'\in A'$. Since $S$ is dense, there is a pair $s,s'$
from $S$ with $s\in A$ and $s'\in A'$. Also, $N$ contains
neighbourhoods $N_s$ and $N_{s'}$ with $N_s\subseteq A$ and
$N_{s'}\subseteq A'$. So $N_s$ and $N_{s'}$ are disjoint. Thus, for
sufficiently large $i$, $x$ and $x'$ fall in distinct elements of the
partition $\S_i$. Since this is true for any pair $x,x'$, it follows
that
$$
\bigcap_{i=1}^\infty \{S\in\S_i: x\in S\}\subseteq\{x\}.
$$
The reverse inclusion is trivial. 
The measurability of all singletons $\{x\}$ follows from the
measurability of $S_x := \bigcup_i
\{S\in \S_i\colon S \cap \{x\} = \phi \}$ and the fact that $\{x\} =
X - S_x$. 
\end{proof}

We shall use Lemma~\ref{lem:topheavy} together with the following result
to show that we can approximate expectations of certain random variables
using a single sample path of the Markov chain.

\begin{lemma}\label{lem:RNderivs}
Let $(X,\B)$ be a measurable space satisfying the conditions of
Lemma~\ref{lem:topheavy}, and let $\S_1,\S_2,\ldots$ be a suitable
sequence of partitions as in that lemma.  Let $\mu$ be a probability
measure defined on this space. Let $f$ be an absolutely integrable
function on $X$. For an event $S$, define
$$
f(S) = \frac{\int_S f \, d\mu}{\mu(S)}.
$$
For each $x\in X$ and $k=1,2,\ldots$, let $S_k(x)$ be the unique
element of $\S_k$ containing $x$. Then for almost all $x$ in $X$,
$$
\lim_{k\to\infty} f(S_k(x)) = f(x).
$$
\end{lemma}

\begin{proof}
Clearly, the signed finite measure $\phi$ defined by
\begin{equation}\label{eq:RDdef1}
\phi(E)=\int_Efd\mu
\end{equation}
is absolutely continuous with respect to $\mu$, and
Equation~\eqref{eq:RDdef1} defines $f$ as the Radon-Nikodym derivative
of $\phi$ with respect to $\mu$. This derivative can also be defined
as
$$
\frac{d\phi}{d\mu}(x) = \lim_{k\to\infty} \frac{\phi(S_k(x))}{\mu(S_k(x))}.
$$
See, for example,~\cite[Section~10.2]{shilov66}.  By the Radon-Nikodym
Theorem~\cite[Theorem~5.5.4, p.~134]{dudley89}, these two expressions
are equal a.e.~($\mu$).
\end{proof}

\begin{proof}{\it (Theorem~\ref{theorem:genpgradmdp}.)}~
From the definitions,
\begin{align}
\nb\eta &= \pi'\nabla P J_\beta \notag \\
        &= \sum_{i=1}^n \sum_{j=1}^n \pi(i) \nabla p_{ij}(\theta)
        J_\beta(j).
\label{eq:vandal}
\end{align}
For every $y$, $\mu$ is absolutely continuous with respect to
the reference measure $\lambda$, hence for any $i$ and $j$ we can write
$$
p_{ij}(\theta) = \int_\Y\int_\U p_{ij}(u)\,
        \frac{d\mu(\theta,y)}{d\lambda}(u) \, d\lambda(u) \, d\nu(i)(y).
$$
Since $\lambda$ and $\nu$ do not depend on $\theta$ and
$d\mu(\theta,y)/d\lambda$ is absolutely integrable, we can
differentiate under the integral to obtain
$$
\nabla p_{ij}(\theta) = \int_\Y\int_\U p_{ij}(u)\,
\nabla \frac{d\mu(\theta,y)}{d\lambda}(u) \, d\lambda(u) \, d\nu(i)(y).
$$
To avoid cluttering the notation, we shall use $\mu$ to denote the
distribution $\mu(\theta,y)$ on $\U$, and $\nu$ to denote the
distribution $\nu(i)$ on $\Y$.  With this notation, we have
$$
\nabla p_{ij}(\theta) = \int_\Y\int_\U p_{ij}\,
  \frac{\nabla \frac{d\mu}{d\lambda}}{\frac{d\mu}{d\lambda}}
  \, d\mu \, d\nu.
$$
Now, let $\rho$ be the probability measure on $\Y\times\U$
generated by $\mu$ and $\nu$. We can write~\eqref{eq:vandal} as
$$
\nb\eta = \sum_{i,j} \pi(i) J_\beta(j)
        \int_{\Y\times\U} p_{ij}
        \frac{\nabla \frac{d\mu}{d\lambda}}{\frac{d\mu}{d\lambda}}
        \, d\rho.
$$
Using the notation of Lemma~\ref{lem:RNderivs}, we define
\begin{align*}
p_{ij}(S) & = \frac{\int_S p_{ij} \, d\rho}{\rho(S)},\\
\nabla(S) & = \frac{1}{\rho(S)}
        \int_S \frac{\nabla\frac{d\mu}{d\lambda}}{\frac{d\mu}{d\lambda}}
        \, d\rho,
\end{align*}
for a measurable set $S\subseteq\Y\times\U$. Notice that, for a
given $i$, $j$, and $S$,
\begin{align*}
p_{ij}(S) & = \Pr\left(X_{t+1}=j \left| X_t=i,\, (y,u)\in
        S\right.\right) \\
\nabla(S) & = \Expect\left(\left.
        \frac{\nabla\frac{d\mu}{d\lambda}}{\frac{d\mu}{d\lambda}}
                \right| X_t=i,\, (Y_t,U_t)\in S\right).
\end{align*}
Let $\S_1,\S_2,\ldots$ be a sequence of partitions of
$\Y\times\U$ as in Lemma~\ref{lem:topheavy}, and let $S_k(y,u)$
denote the element of $\S_k$ containing $(y,u)$.  Using
Lemma~\ref{lem:RNderivs}, we have
\begin{align*}
\int_{\Y\times\U} p_{ij} \,
        \frac{\nabla\frac{d\mu}{d\lambda}}{\frac{d\mu}{d\lambda}} \,
        d\rho
 &=     \int_{\Y\times\U} \lim_{k\to\infty}
        p_{ij}\left(S_k(y,u)\right) \, \nabla\left(S_k(y,u)\right) \,
        d\rho(y,u) \\
 &=     \lim_{k\to\infty} \sum_{S\in\S_k} \int_S
        p_{ij}(S) \, \nabla(S) \, d\rho,
\end{align*}
where we have used Assumption~\ref{ass:densmubound} and the Lebesgue
dominated convergence theorem to interchange the integral and the
limit. Hence,
\begin{align*}
\nb\eta
        &= \lim_{k\to\infty} \sum_{i,j} \sum_{S\in\S_k}
                \pi(i) \rho(S) p_{ij}(S) J_\beta(j) \nabla(S) \\
        &= \lim_{k\to\infty} \sum_{i,j,S} \Pr(X_t=i) \Pr((Y_t,U_t)\in S)
                \Pr\left(X_{t+1}=j\left|X_t=i,\, (Y_t,U_t)\in S\right.\right)
                \\
        &\qquad \Expect\left(J(t+1)| X_{t+1}=j\right)
                \Expect\left(\left.
                \frac{\nabla\frac{d\mu}{d\lambda}}{\frac{d\mu}{d\lambda}}
                \right| X_t=i,\, (Y_t,U_t)\in S\right) \\
        &= \lim_{k\to\infty} \sum_{i,j,S} \Expect\left[
                \chi_i(X_t) \chi_S(Y_t,U_t) \chi_j(X_{t+1}) J(t+1)
                \frac{\nabla\frac{d\mu}{d\lambda}}{\frac{d\mu}{d\lambda}}
                \right],
\end{align*}
where probabilities and expectations are with respect to the
stationary distribution $\pi$ of $X_t$, and the distributions on $Y_t,
U_t$. Now, the random process inside the expectation is asymptotically
stationary and ergodic. From the ergodic theorem, we have (almost
surely)
$$
\nb\eta
        = \lim_{k\to\infty} \lim_{T\to\infty} \frac{1}{T} \sum_{i,j,S}
                \sum_{t=0}^{T-1} \chi_i(X_t) \chi_S(Y_t,U_t) \chi_j(X_{t+1})
                J(t+1)
                \frac{\nabla\frac{d\mu}{d\lambda}}{\frac{d\mu}{d\lambda}}.
$$
It is easy to see that the double limit also exists when the order is
reversed, so
\begin{align*}
\nb\eta
        &= \lim_{T\to\infty} \frac{1}{T} \sum_{t=0}^{T-1}
                \lim_{k\to\infty} \sum_{i,j,S}
                \chi_i(X_t) \chi_S(Y_t, U_t) \chi_j(X_{t+1}) J(t+1)
                \frac{\nabla\frac{d\mu}{d\lambda}}{\frac{d\mu}{d\lambda}} \\
        &= \lim_{T\to\infty} \frac{1}{T} \sum_{t=0}^{T-1}
                \frac{\nabla\frac{d\mu(\theta,Y_t)}{d\lambda}(U_t)}
                     {\frac{d\mu(\theta,Y_t)}{d\lambda}(U_t)} J(t+1).
\end{align*}
The same argument as in the proof of Theorem~\ref{theorem:gradmdp}
shows that the tails of $J(t+1)$ can be ignored when
$$
\left|\frac{\nabla\frac{d\mu(\theta,Y_t)}{d\lambda}(U_t)}
                     {\frac{d\mu(\theta,Y_t)}{d\lambda}(U_t)}\right|
$$
and $|r(X_t)|$ are uniformly bounded. It follows that
$\Delta_T\to\pi'\nabla P J_\beta$ w.p.1, as required.
\end{proof}

\bibliographystyle{theapa}
\bibliography{bib}

\end{document}